%% file: document.tex
\documentclass[11pt,twoside,letterpaper]{article} 
\usepackage{times,fancyhdr}

\usepackage[margin=1.9cm]{geometry}
\usepackage{parskip}
\usepackage{times} 
\usepackage{amssymb,amsmath}
\usepackage{amsthm}
\usepackage{mathptmx}
\usepackage{algorithms/algorithm}
\usepackage{algorithms/algorithmic}

\usepackage{graphicx}
\usepackage{sidecap}
\usepackage{subfigure}
\usepackage{wrapfig}

\def\tT{{\mbox{\tiny{T}}}}
\newcommand{\bitem}{\begin{itemize}}
\newcommand{\eitem}{\end{itemize}}

\newcommand{\bpm}{\begin{pmatrix}}
\newcommand{\epm}{\end{pmatrix}}

\newcommand{\bq}{\begin{equation}}
\newcommand{\eq}{\end{equation}}

\newcommand{\D}{\mathbb{D}iv \;}

\newcommand{\C}{\mathcal{C}}
\newcommand{\R}{\mathcal{R}}

\let\abs=\envert

\let\norm=\enVert

\let\inprod=\inProd

\newtheorem{theorem}{Theorem}

\newtheorem{coro}[theorem]{Corollary}

\newtheorem{prop}[theorem]{Proposition}


\makeatletter 
\def\cleardoublepage{\clearpage\if@twoside \ifodd\c@page\else%
    \hbox{}%
    \thispagestyle{empty}%
    \newpage%
    \if@twocolumn\hbox{}\newpage\fi\fi\fi} 
\makeatother

\renewcommand{\thesection}{\arabic{section}.}
\renewcommand{\thesubsection}{\thesection\arabic{subsection}.}

\def\figurename{Figure}
\makeatletter
\renewcommand{\fnum@figure}[1]{\figurename~\thefigure.}
\makeatother

\def\tablename{Table}
\makeatletter
\renewcommand{\fnum@table}[1]{\tablename~\thetable.}
\makeatother

\begin{document}
\title{
{
\vskip 0.45in
\bfseries\scshape 
Modern Convex Optimization to Medical Image Analysis}} \author{\bfseries\itshape
Jing Yuan\thanks{E-mail address: jyuan@xidian.edu.cn}\\
School of Mathematics and Statistics, Xidian University, Xi'an,
China \\
\bfseries\itshape Aaron Fenster\thanks{E-mail address: afenster@robarts.ca}\\
Robarts Research Institute, Schulich Medical School, Western
University, Canada
}
\date{}
\maketitle
\thispagestyle{empty}
\setcounter{page}{1}
\thispagestyle{fancy}
\fancyhead{}
\fancyhead[L]{In: Book Title \\ 
Editor: Editor Name, pp. {\thepage-\pageref{lastpage-01}}} 
\fancyhead[R]{ISBN 0000000000  \\
\copyright~2007 Nova Science Publishers, Inc.}
\fancyfoot{}
\renewcommand{\headrulewidth}{0pt}

\begin{abstract}
Recently, diagnosis, therapy and monitoring of human
diseases involve a variety of imaging modalities, such as magnetic resonance
imaging(MRI), computed tomography(CT), Ultrasound(US) and Positron-emission
tomography(PET) as well as a variety of modern optical techniques.
Over the past two decade, it has been recognized that advanced image processing
techniques provide valuable information to physicians for diagnosis, image
guided therapy and surgery, and monitoring of the treated organ to the therapy.
Many researchers and companies have invested significant efforts
in the developments of advanced medical image analysis methods; especially in the two core
studies of medical image segmentation and registration,
segmentations of organs and lesions are used to quantify volumes and shapes used in diagnosis and monitoring treatment;
registration of multimodality images of organs improves detection, diagnosis and
staging of diseases as well as image-guided surgery and therapy, 
registration of
images obtained from the same modality are used to monitor progression of
therapy. 
These challenging
clinical-motivated applications introduce novel and sophisticated mathematical
problems which stimulate developments of advanced optimization and computing
methods, especially convex optimization attaining optimum in a global
sense, hence, bring an enormous spread of research topics for recent
computational medical image analysis. Particularly, distinct from the usual
image processing, most medical images have a big volume of acquired data,
often in 3D or 4D (3D + t) along with great noises or incomplete image
information, and form the challenging large-scale
optimization problems; how to process such poor 'big data' of medical images
efficiently and solve the corresponding optimization problems robustly are the
key factors of modern medical image analysis.
\end{abstract}


\vspace{.08in} 

\pagestyle{fancy}
\fancyhead{}
\fancyhead[EC]{J. Yuan, A. Fenster}
\fancyhead[EL,OR]{\thepage}
\fancyhead[OC]{Medical Image Analysis Meets Modern Convex Optimization}
\fancyfoot{}
\renewcommand\headrulewidth{0.5pt} 

\section{Introduction}

Recently, diagnosis, therapy and monitoring of human
diseases involve a variety of imaging modalities, such as magnetic resonance
imaging(MRI), computed tomography(CT), Ultrasound(US) and Positron-emission
tomography(PET) as well as a variety of modern optical techniques.
Over the past two decade, it has been recognized that advanced image processing
techniques provide valuable information to physicians for diagnosis, image
guided therapy and surgery, and monitoring of the treated organ to the therapy.
Many researchers and companies have invested significant efforts
in the developments of advanced medical image analysis methods; especially in the two core
studies of medical image segmentation and registration,
segmentations of organs and lesions are used to quantify volumes and shapes used in diagnosis and monitoring treatment;
registration of multimodality images of organs improves detection, diagnosis and
staging of diseases as well as image-guided surgery and therapy, 
registration of
images obtained from the same modality are used to monitor progression of
therapy. 
In this work, we focus on these two most challenging
problems of medical image analysis and show recent progresses in
developing efficient and robust computational tools by modern convex
optimization.

Thanks to a series of pioneering works
\cite{Nikolova2006,ChambolleP11,Yuan2010} during recent ten years, convex 
optimization was developed as a powerful tool to analyze and solve most
variational problems of image processing, computer vision and machine learning efficiently.
For example, the total-variation-based image denoising~\cite{rudin1992nonlinear,goldstein2014fast}
\[
\min_u \;\; \int D(u-f) \, dx \, + \, \alpha \int \abs{\nabla u} \, dx \, ,
\]
where $D(\cdot)$ is a convex penalty function, e.g. $L_1$ or $L_2$ norm; the $L_1$-normed sparse image reconstruction~\cite{Beck2009A}
\[
\min_u \;\; \int D(Au-f) \, dx \, + \, \alpha \int \abs{u} \, dx \, ,
\] 
where $A$ is some linear operator; and many other problems which are initially nonconvex but can be finally solved by convex optimization, such as  
the spatially continuous min-cut model for image segmentation~\cite{Nikolova2006,Yuan2010}
\bq \label{eq:minc}
\min_{u(x) \in \{0,1\}} \;\; \int u(x) \, C(x)\, dx \, + \,  \alpha \int \abs{\nabla u} \, dx \, ,
\eq
for which its binary constraint can be relaxed as $u(x) \in [0,1]$, hence results in a convex optimzation problem \cite{Nikolova2006}. 

In this paper, we consider the
optimization problems of medical image segmentation and registration as the minimization of a
finite sum of convex function terms:
\bq \label{eq:pmodel}
\min_u \;\; f_1(u) \, + \, \ldots \, + \, f_n(u) \; ,
\eq
which actually includes the convex constrained optimization problem as one special case such that 
the convex constraint set $\C$ on the unknown function $u(x) \in \C$ 
can be reformulated by adding its convex characteristic function 
\[
\chi_{\C}(u) \, := \, \left\{ \begin{array}{ll}  0 \, , &
\text{$x \in \C$} \\ +\infty \,  , & \text{$x\in \C$} \end{array} \right. \, .
\]
into the energy function of \eqref{eq:pmodel}.

Given the very high dimension of the solution $u$, which is the usual case of
medical image analysis where the input image volume often includes over millions of
pixels, the iterative first-order gradient-descent schemes play the central role in builing
up a practical algorithmic implementation, which typically has a affordable
computational cost per iteration along with proved iteration complexity.
In this perspective, the duality of each convex function term $f_i(u) =
\inprod{u,p_i} -f_i^*(p_i)$ provides one most
powerful tool in both analyzing and developing such first-order 
iterative algorithms, where the introduced new dual variable $p_i$ for each
function term $f_i$ just represents the first-order gradient of $f_i(u)$
implicitly; it brings two equivalent optimization models, a.k.a.
the \emph{primal-dual model} 
\bq \label{eq:pdmodel}
\min_u \max_p \; \; \underbrace{\inprod{p_1 + \ldots + p_n, u} \, - \,
f_1^*(p_1) \, - \, \ldots \, - \, f_n^*(p_n)}_{\text{Lagrangian function } L(u,
p)}\;
\eq
and the \emph{dual model} 
\bq \label{eq:dmodel}
\left. \begin{array}{ll} \max_p \; \; &\,  - \, f_1^*(p_1) \, - \, \ldots
\, - \, f_n^*(p_n)\, 
\\ \text{s.t.  } \,  & \, p_1 + \ldots + p_n \, = \, 0
\end{array}
\right. \, 
\eq
to the studied convex minimization problem \eqref{eq:pmodel}. 

Comparing with the traditional first-order gradient-descent algorithms which
directly evaluate the gradient of each function term at each iteration and improve the 
approximation of optimum iteratively, 
the dual model \eqref{eq:dmodel} provides another expression to analyze the
original convex optimization model \eqref{eq:pmodel} and delivers a novel point
of view to design new first-order iterative algorithms, where the optimum $u^*$
of \eqref{eq:pmodel} just works as the optimal multiplier to the linear equality
constraint as demonstrated in the Lagrangian function $L(u,p)$ of the primal-dual 
model \eqref{eq:pdmodel} (see more details in
Sec.~\ref{sec:optim}). 
In practice, such dual formulation based approach enjoys great advantages in
both mathematical analysis and algorithmic design:
a. each function term $f_i(p_i)$ of its energy function depends solely on an
independent variable $p_i$, which naturally leads to an efficient splitting
scheme to tackle the optimization problem in a simple separate-and-conquer way,
or a stochastic descent scheme with low iteration-cost; {b.
a unified algorithmic framework to compute the optimum multiplier $u^*$ can be
developed by the \emph{augmented Lagrangian method} (ALM)}, which involves two
sequential steps at each iteration:
\begin{align} \label{eq:ALM}
p^{k+1} \, := & \, \arg\max_p \, L(u^k,p) \, - \, \frac{c^k}{2}\norm{p_1
+ \ldots + p_n}^2 \, , \\
u^{k+1} \, = & \, u^k \, -
\, c^k(p_1^{k+1} + \ldots + p_n^{k+1}) \, ,
\end{align}
with capable of setting up high-performance parallel implementations under the
same numerical perspective;
{c. the equivalent dual model in \eqref{eq:dmodel} additionally
brings new insights to facilitate analyzing its original 
model \eqref{eq:pmodel} and discovers close connections from distinct
optimization topics (see Sec. \ref{sec:seg} and \ref{sec:reg} for details).} 

\begin{figure}[ht!]
{\subfigure[]{\includegraphics[height=3.2cm]{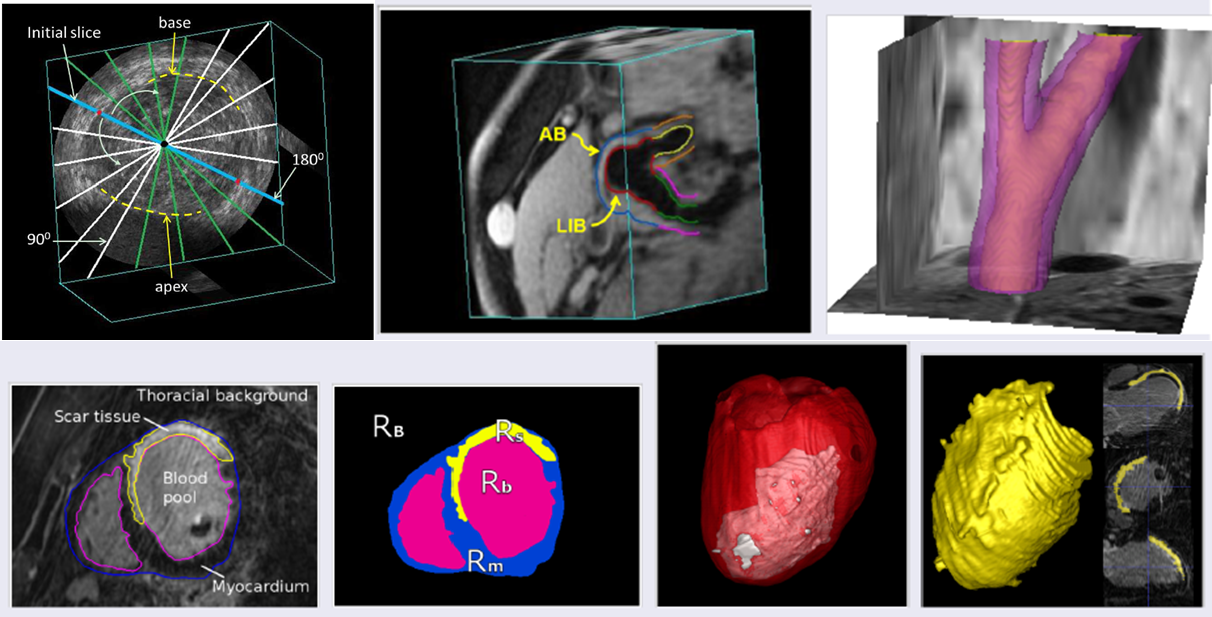}}
\subfigure[]{\includegraphics[height=3.2cm]{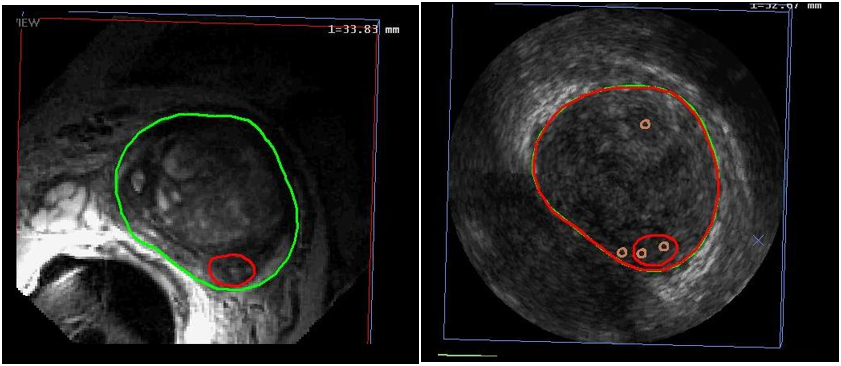}}
\subfigure[]{\includegraphics[height=2.2cm]{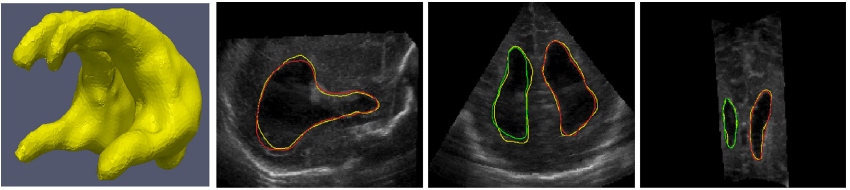}}
\subfigure[]{\includegraphics[height=2.2cm]{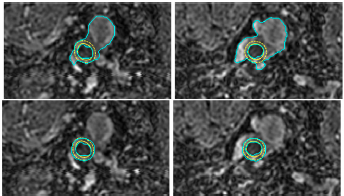}}
}
{\caption{ \textit{ (a) Medical image segmentation examples of 3D
prostate ultrasound with shape symmetry
prior~\cite{yuan2013efficient,qiu2014prostate}, 
carotid artery AB-LIB MRI
with linear overlaid prior~\cite{Ukwatta2012,ukwatta20133}, 
cardiac scar tissues
from LEC cardiac MR images with partial region-order
prior~\cite{rajchl2012fast,rajchl2014interactive}.
(b) Multi-modality nonrigid image registration of 3D prostate MR/TRUS images.
(c) Extracting neonatal brain ventricles from 3D ultrasound
image using globally optimized multi-region level-sets~\cite{QiuLS2015}. 
(d) Segmenting lumen-outer walls from femoral arteries MR
images~\cite{ukwatta20133} 
without (top row) and with (bottom row) the spatial region consistency.}}\label{fig:illus1}}
\vspace{-0.3cm}
\end{figure}

For the studies of medical image segmentation and non-rigid registration, our
studies showed that such dual optimization methods largely improved
efficiency and accuracy in numerical practices and
reduced manual efforts and intra- and inter-observer variabilities. 
Meanwhile, the dual optimization method can easily incorporate
prior information and constraints into the optimization models, which
perfectly reduces bias from low image quality and incomplete imaging
information often appearing in most medical image modalities. Such prior
information includes anatomical knowledge and machine-learned
image features, for example: linear or partial region orders, shape symmetry and
compactness, region volume-preserving and spatial consistency etc., and were
successfully applied to many different applications, e.g. the
segmentation of carotid artery adventitia boundary(AB) and lumen-intima
boundary(LIB) from T1-weighted
black-blood carotid magnetic resonance (MR)
images~\cite{Ukwatta2012,ukwatta20133}, 
segmenting scar tissues from
Late-Enhancement Cardiac MR
Images~\cite{rajchl2012fast,rajchl2014interactive,Baxter2017Directed}, prostate zonal segmentation from T2-MRIs, 
registration of 3D
TRUS image to MR image~\cite{QY2016}, 
3D prostate ultrasound image
segmentation~\cite{yuan2013efficient,qiu2014prostate,Ukwatta2015120},
co-segmenting lung pulmonary $^1$H and hyperpolarized $^3$He MRIs~\cite{Guo2015Globally},
3D non-rigid registration of prostate MRI-TRUS~\cite{sun2015a},
4D spatial-temporal deformable registration of newborn brain ultrasound images for monitoring pre-term
neonatal cerebral ventricles~\cite{qy2015miccai,qiuyj2017} etc.

\vspace{0.3cm}
{\bf Organization: } 
The following contents of this work is organized in three parts:
In Sec.~\ref{sec:optim}, we introduce the main theories and algorithmic scheme used in convex optimization
under a unified dual optimization framework; the dual optimization
approach sets up new equivalent optimization formulations to the studied convex optimization model
and derives a new unified multiplier-based algorithmic framework; in addition, some
applications of image processing are presented as examples.
In Sec.~\ref{sec:seg}, we study the clinical-motivated
applications of medical image segmentation and show the introduced dual optimization approach
can easily integrate various prior information, which largely improves
the accuracy and robustness of optimization solutions, 
into the ALM-based optimization algorithms with
much less efforts than tackling the original convex optimization formulations directly.
In Sec.~\ref{sec:reg}, we demonstrate that the introduced dual optimization approach
can be explored to solve the challenging non-rigid image registration problem, especially with
the additional volume-preserving and temporal consistency prior; experiment results from real
clinical applications showed its great performance in practice.

\input{optim.tex}

\input{seg.tex}

\input{reg.tex}


\bibliographystyle{plain}
\bibliography{refs,cmfls_mls}

\label{lastpage-01}

\end{document}

%% file: optim.tex
\section{Convex Optimization and Dual Optimization Method}
\label{sec:optim}

In this section, we consider the convex optimization problem \eqref{eq:pmodel} of minimizing the
sum of multiple convex function terms, 
which generalizes a big spectrum of convex optimization models
including the convex constrained optimization problem for which the convex constraint set $\C$ on $u(x) \in \C$
can be well imposed by adding its characteristic function $\chi_{\C}(u)$ as one function term in \eqref{eq:pmodel}.

As one powerful tool to analyze convex functions, the duality
of a convex function was developed \cite{MR0274683} and largely exploited in
designing fast convex optimization algorithms \cite{ChambolleP11} to
image processing recently, such that each convex function $f_i(u)$ of
\eqref{eq:pmodel} can be equally represented by
\bq \label{eq:conj}
f_i(u) \, = \, \max_{p_i} \, \inprod{u,p_i} -f_i^*(p_i)
\eq
where $f_i^*(p_i)$ is the corresponding conjugate function of $f_i(u)$ and
$\inprod{u, p_i}$ is the inner product of $u$ and the dual variable $p_i$.

Therefore, by simple computation, we have the following result
\begin{prop} \label{prop:01}
The convex optimization problem \eqref{eq:pmodel} is mathematically equivalent
to the linear equality constrained convex optimization problem:
\bq \label{dual-model}
\max_p \; - f_1^*(p_1) - \ldots - f_n^*(p_n) \, , \quad
\text{s.t.} \; p_1 + .. + p_n = 0\, ;
\eq
i.e. the dual optimization model~\eqref{eq:dmodel}.
\end{prop}
Its proof comes from the following facts: first, summing up each conjugate expression \eqref{eq:conj} of the convex 
function $f_i(u)$, $i=1 \ldots n$, results in a Lagrangian formulation
\bq \label{pd-model}
\max_p \min_u \; L(u,p) \, := \, \inprod{p_1 + \ldots + p_n, u} \, - \,
f_1^*(p_1) \, - \, \ldots \, - \, f_n^*(p_n) \, .
\eq
Given the convexity of $L(u,p)$ on each variable of $u$ and $p$, the minimization and 
maximization procedures of \eqref{pd-model} are actually interchangeable~\cite{citeulike1859441}.
Then, the minimization of~\eqref{pd-model} over $u$ leads to the vanishing of the 
inner product term $\inprod{p_1 + \ldots + p_n, u}$ and the corresponding linear equality constraint
$
p_1 + .. + p_n  = 0
$.
This, hence, gives rise to the dual model \eqref{dual-model}.

Actually, for each convex function $f_i(u)$, the optimum of $p_i$ for its dual expression \eqref{eq:conj} is nothing but
its corresponding gradient or subgradient at $u$; therefore, the linear equality constraint $p_1 + .. + p_n  = 0$ for the dual model
\eqref{dual-model} exactly represents the first-order optimal condition to the studied convex optimization problem \eqref{eq:pmodel}, i.e.
the sum of all gradients or subgradients vanishes
\[
\partial f_1(u) \, + \,  \ldots \, + \, \partial f_n(u) \, = \, 0 \, .
\]

Additionally, we can further conclude that
\begin{coro} \label{coro:01}
For the dual optimization problem \eqref{dual-model}, the optimum multiplier $u^*$ to its linear equality constraint $p_1 + .. + p_n  = 0$ is
just the minimum of the original convex optimization problem \eqref{eq:pmodel}.
\end{coro}

This is clear from the above proof to Prop.~\ref{prop:01}. Especially, this result establishes the basis of a novel algorithmic framework 
to a wide spectrum of convex optimization problems using the classical augmenmted Lagrangian method (ALM)~\cite{citeulike1859441,Rockafellar1976} 
(see Sec.~\ref{sec:dom} for more details).

Especially, each function term $f_i^*(p_i)$, $i=1 \ldots n$, in the energy function of 
the dual model \eqref{dual-model} solely depends on an independent variable $p_i$ which is loosely correlated to the
other variables by the linear equality constraint $p_1 + .. + p_n  = 0$. This is in contrast to
its original optimization model \eqref{eq:pmodel} whose energy function terms are interracted with each with the
common unknown variable $u$. This provides a big advantage in develop splitting optimization algorithms, as shown in Sec.~\ref{sec:dom},
to tackle the underlying convex optimization problem, particularly at a large scale.

\subsection{Dual Optimization Method}
\label{sec:dom}

Now we consider the linearly constrained convex optimization problem \eqref{dual-model}, i.e. the dual model, and the corollary~\ref{coro:01}, such that 
the energy function $L(u,p)$ of the primal-dual model \eqref{pd-model} is exactly the Lagrangian function of the linearly
constrained dual model \eqref{dual-model}.
Hence, the classical augmented Lagrangian method (ALM)~\cite{citeulike1859441,Rockafellar1976} provides an optimization framework to develop 
the corresponding algorithmic scheme. 

For this, we define the associate augmented Lagrangian function 
\bq \label{eq:aug-func}
 L_c(u,p) \, :=  \underbrace{- \,
f_1^*(p_1) \, - \, \ldots \, - \, f_n^*(p_n) \, + \, \inprod{p_1 + \ldots + p_n, u}}_{L(u,p)} \,
- \, \frac{c}{2}\norm{p_1 + \ldots + p_n}^2\, ,
\eq
where $c$ is the positive parameter.

\begin{algorithm}[!h]
\caption{Augmented Lagrangian Method Based Algorithm\label{algd}} 
Initialize $u^0$ and $(p_1^0, \ldots, p_n^0)$, for each iteration $k$ we explore the following two steps
\begin{itemize}
  \item fix $u^k$, compute $p^{k+1}$:
  \bq \label{alg:stepp}
  (p_1^{k+1}, \ldots, p_{n}^{k+1}) \, = \, \arg\max_p \, L_{c^k}(u^k, p) \, ;
  \eq
  \item fix $p^{k+1}$, then update $u^{k+1}$ by
  \bq 
  u^{k+1} \, :=\, u^k - c^k\big(p_1^{k+1} + p_2^{k+1} + .. + p_n^{k+1}\big) \label{alg:stepu} 
  \eq
\end{itemize}
\end{algorithm}

Then, an ALM-based algorithm can be developed as shown in Alg.~\ref{algd}, which explores two consecutive optimization steps \eqref{alg:stepp} and \eqref{alg:stepu}
over $p$ and $u$ correspondingly. The convergence of such ALM-based can be proved to obtain a linear rate of $O(1/N)$ \cite{He2012On}.

Clearly, at each iteration, the main computing load is from the first optimization step \eqref{alg:stepp}. In practice, the optimization 
sub-problem \eqref{alg:stepp} is often solved by tackling each $p_i$, $i=1 \ldots n$, 
separately. More specifically, this can be implemented either in parallel such that
\begin{align}
p_1^{k+1} \, := & \, \arg\max \Big\{ -f_1^*(p_1) - \frac{c^k}{2}\norm{p_1 + p_2^k + .. + p_n^k - \frac{u^k}{c^k}}^2 \Big\} \label{alg:pstepp1}\\
& ... ...\nonumber \\
p_n^{k+1} \, := & \, \arg\max \Big\{ -f_n^*(p_n) - \frac{c^k}{2}\norm{p_1^{k} + p_2^{k} + .. + p_n - \frac{u^k}{c^k}}^2 \Big\} \, ,\label{alg:psteppn}
\end{align}
or in a sequential way such that
\begin{align}
p_1^{k+1} \, := & \, \arg\max \Big\{ -f_1^*(p_1) - \frac{c^k}{2}\norm{p_1 + p_2^k + .. + p_n^k - \frac{u^k}{c^k}}^2 \Big\} \label{alg:ssteps1}\\
& ... ...\nonumber \\
p_n^{k+1} \, := & \, \arg\max \Big\{ -f_n^*(p_n) - \frac{c^k}{2}\norm{p_1^{k+1} + p_2^{k+1} + .. + p_n - \frac{u^k}{c^k}}^2 \Big\} \, . \label{alg:sstepsn}
\end{align}

Particularly, every optimization sub-problem of \eqref{alg:pstepp1}-\eqref{alg:psteppn} or \eqref{alg:ssteps1}-\eqref{alg:sstepsn} is approximately solved by one step of
gradient-descent in order to alleviate computational complexities of each iteration.

\subsection{Some Applications to Image Processing}

\noindent{\bf Total-Variation-Based Image Denoising:} 
For the total-variation-based image denoising, it can be formulated as the following convex optimization problem
\bq \label{eq:img-tv}
\min_u \; D(u-f) \, + \, \alpha \int_{\Omega} \abs{\nabla u}\, dx \, , 
\eq
where $D(\cdot)$ is some convex data fidelity function, typically some convex function, for example, 
the L2-norm such that $D(\cdot)=\frac{1}{2}\norm{\cdot}_2^2$ gives rise to the application of TV-L2 image denoising, i.e.
\bq
\min_u \;  \frac{1}{2}\int_{\Omega} \abs{u-f}^2\, dx \, + \, \alpha \int_{\Omega} \abs{\nabla u}\, dx \, , 
\eq
and the L1-norm such that $D(\cdot)=\norm{\cdot}_{1}$ results in the application of TV-L1 image denoising:
\bq
\min_u \;  \int_{\Omega} \abs{u-f}\, dx \, + \, \alpha \int_{\Omega} \abs{\nabla u}\, dx \, .
\eq

Let $D^*(q)$ be the conjugate of the convex function $D(\cdot)$ such that 
\[
D(u-f) \, = \, \max_q \, \inprod{q, u-f} \, - \, D^*(q) \, ,
\]
where, for the case of L2-norm, we have
\bq \label{eq:l2-tvfunc}
D^*(q) \, = \, \frac{1}{2}\norm{q}_2^2\, = \, \frac{1}{2}\int_{\Omega}q^2\, dx \, ,
\eq
and for the case of L1-norm, $D^*(q)$ is just the indicator function of the convex set such that: 
\[
D^*(q) = \left\{ \begin{array}{rl}
       0 \,, & \qquad \text{if $q(x) \leq 1$} \\     
       +\infty\, , & \qquad \text{otherwise} 
    \end{array} \right. \, .
\]

Given the dual formulation of the total-variation function~\cite{citeulike3001108}
\bq \label{eq:dtvreg}
\alpha \int_{\Omega}\abs{\nabla u}\, dx \,= \, \max_{\abs{p(x)} \leq \alpha} \, \int_{\Omega} u \,\, \D p \, dx\, ,
\eq
we can easily rewrite the total-variation-based image denoising problem \eqref{eq:img-tv} by
\bq \label{eq:pd-img-tv}
\max_{q,p} \min_u \; L(u, p, q)\, :=\, -\inprod{q, f} \, - \, D^*(q) \, + \, \inprod{u, q \, + \, \D p}\, ,
\eq
i.e. the corresponding primal-dual model.
Minimizing the Lagrangian function $L(u,p)$ of the primal-dual model \eqref{eq:pd-img-tv} over $u$, we can derive
the equivalent dual optimization model to the total-variation-based image denoising problem \eqref{eq:img-tv}
\bq \label{eq:img-tv-dual}
\max_{q,p} \; -\inprod{q, f} \, - \, D^*(q) \, , \qquad \text{s.t.} \;\; q \, + \, \D p \, = \, 0 \, .
\eq
In view of Coro.~\ref{coro:01}, the optimum $u^*$ to the original image denoising optimization problem \eqref{eq:img-tv} 
is just the optimal multiplier to the above linear equality constraint of \eqref{eq:img-tv-dual} 
under the perspective of the dual formulation~\eqref{eq:img-tv-dual}.

\begin{figure}[ht!]
{\subfigure[]{
\includegraphics[height=2.7cm]{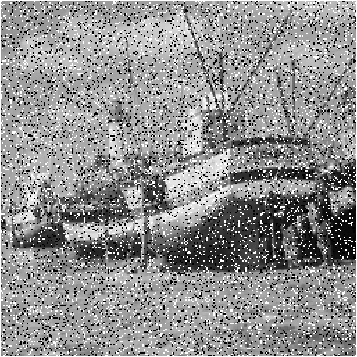}
\includegraphics[height=2.7cm]{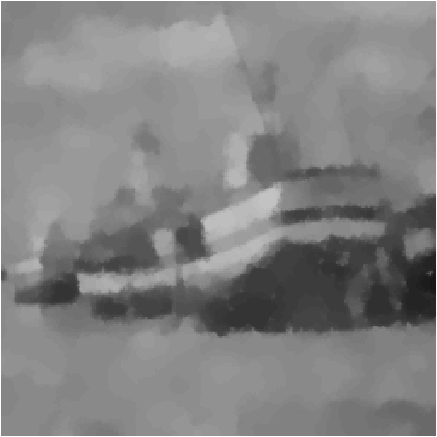}
\includegraphics[height=2.7cm]{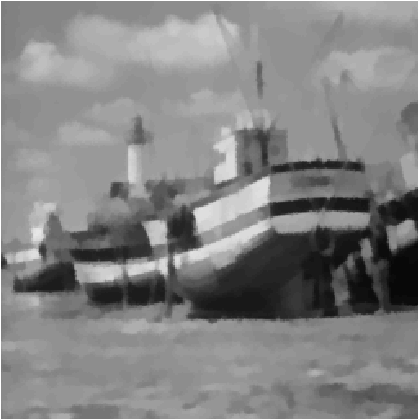}
}
\subfigure[]{
\includegraphics[height=2.7cm]{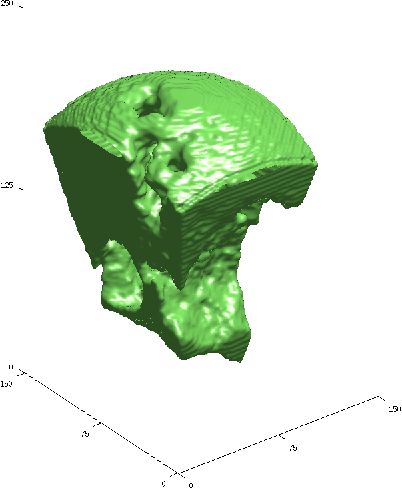}
}
\subfigure[]{
\includegraphics[height=2.7cm]{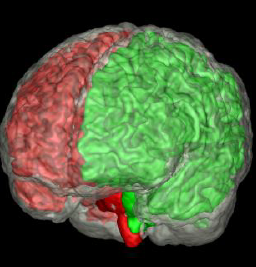}
}
}
{\caption{ \textit{ (a) Example of total-variation-based image denoising computed by the dual formulation \eqref{eq:img-tv-dual} based Alg.~\ref{img-tv-algd}: left:
the input image with noises, middle: its denoised image using L2-norm as its data fidelity term,
right: its denoised image using L1-norm as its data fidelity term;
(b) foreground-background segmentation result of a 3D cardiac ultrasound image, computed by
the continuous max-flow model \eqref{eq:seg-mf} based Alg.~\ref{seg-algd} (code is available at \cite{Cut-code}).  
(c) multiphase segmentation of a 3D brain CT image, computed by
the continuous max-flow model \eqref{eq:seg-mf} based Alg.~\ref{seg-algd} (code is available at \cite{Potts-code}).
}}\label{fig:illus2}}
\vspace{-0.3cm}
\end{figure}

In addition, we define the augmented Lagrangian function according to \eqref{eq:pd-img-tv}
\bq \label{eq:augL-img-tv}
L_c(u,p,q) \, = \, \underbrace{ -\inprod{q, f} \, - \, D^*(q) \, + \, \inprod{u, q \, + \, \D p}}_{L(u,p,q)} \, - \, \frac{c}{2}\norm{q \, + \, \D p}^2 \, ;
\eq
similar as the augmented Lagrangian method (ALM) based algorithm~\ref{algd}, we have the ALM based image denoising algorithm (Alg.~\ref{img-tv-algd}).
Fig.~\ref{fig:illus2} (a) shows an illustration for the dual model \eqref{eq:img-tv-dual} based image denoising, where L2-norm and L1-norm are used as 
data fidelity functions respectively.

\begin{algorithm}[!h]
\caption{ALM Based Total-Variation Image Denoising Algorithm\label{img-tv-algd}} 
Initialize $u^0$ and $(q^0, p^0)$, for each iteration $k$ we explore the following two steps
\begin{itemize}
  \item fix $u^k$, compute $(q^{k+1},p^{k+1})$:
  \bq \label{alg:img-tv-stepp}
  (q^{k+1}, p^{k+1}) \, = \, \arg\max_{q,p} \, L_{c^k}(u^k, p, q) \, ;
  \eq
  \item fix $(q^{k+1},p^{k+1})$, then update $u^{k+1}$ by
  \bq 
  u^{k+1} \, :=\, u^k - c^k\big(q^{k+1} \, + \, \D p^{k+1}\big) \label{alg:img-tvstepu} 
  \eq
\end{itemize}
\end{algorithm}

\vspace{0.5cm}
\noindent{\bf Min-Cut-Based Image Segmentation:}
During the last decades, the min-cut model was developed to become one of the most successful model for 
image segmentation~\cite{Boykov01anexperimental,Boykov01fastapproximate}, which has been well studied in the discrete 
graph setting and can be efficiently solved by the scheme of maximizing flows. In fact, such min-cut model can be
also formulated in a spatially continuous setting, i.e.
the \emph{spatially continuous min-cut problem}~\cite{Nikolova2006}:
\bq \label{eq:seg-model}
\min_{u(x) \in \{0,1\}} \;
 \int_{\Omega} \big\{ \big(1-u \big)C_t  + u \, C_s \big\}(x) \, dx + \alpha \int_{\Omega}\abs{\nabla u}\, dx \, ,
\eq
where $C_s(x)$ and $C_t(x)$ are the cost functions such that, for each pixel $x \in \Omega$, $C_s(x)$ and $C_t(x)$ give
the costs to label $x$ as 'foregound' and 'background' respectively. The optimum $u^*(x)$ to the combinatorial optimization
problem~\eqref{eq:seg-model} defines the optimal foreground segmentation region $S$ such that $u^*(x)=1$ for any $x \in S$, and 
the background segmentation region $\Omega\backslash S$ otherwise. 

Chan et al~\cite{Nikolova2006} proved that the challenging non-convex combinatorial optimization problem~\eqref{eq:seg-model}
can be solved globally by computing its convex relaxation model
\bq \label{eq:seg-cr}
\min_{u(x) \in [0,1]} \;
 \int_{\Omega} \big\{ \big(1-u \big)C_s  + u \, C_t \big\} \, dx + \alpha \int_{\Omega}\abs{\nabla u}\, dx \, ,
\eq
while threshholding the optimum of \eqref{eq:seg-cr} with any parameter $\beta \in (0,1)$.

Yuan et al~\cite{Yuan2010,ybtb2014} proposed that the convex relaxed min-cut model \eqref{eq:seg-cr} can be equivalently reformulated
by its dual model, i.e. the \emph{continuous max-flow model}:
\begin{align}
\max_{p_s, p_t, p} \; & \int_{\Omega} p_s(x) \, dx \,  \label{eq:seg-mf} \\
\textbf{s.t.} \;\;& \abs{p(x)}   \leq \alpha \, ,  \quad 
 p_s(x) \leq  C_s(x) \, , \quad
p_t(x) \leq  C_t(x) \,  ;  \label{eq:seg-mf-fc1}\\
& \Big(\D p - p_t + p_s \Big)(x)
\, = \, 0 \,  \,  . \label{eq:seg-mf-fc2}
\end{align}
To see this, we notice that the energy function term $\int_{\Omega} \big\{ \big(1-u \big)C_s  + u \, C_t \big\} \, dx$ in \eqref{eq:seg-cr},
along with the convex constraint $u(x) \in [0,1]$, can be equally expressed by
\[
\max_{p_s, p_t} \; \big(1-u(x) \big) p_t(x) \, + \, u(x) \, p_s(x)\, , \quad \text{s.t.} \; 
p_s(x) \leq  C_s(x) \, , \; p_t(x) \leq  C_t(x) \, ,
\]
and the optimum $u^*(x)$ to the convex relaxed min-cut model \eqref{eq:seg-cr} is exactly the optimum multiplier to
the linear equality constraint \eqref{eq:seg-mf-fc2}
(see \cite{Yuan2010,ybtb2014} for more details).

Correspondingly, we define the augmented Lagrangian function
\bq \label{eq:seg-auglag}
L_c(u, p_s,p_t, p) \, = \,\int_{\Omega} \, p_s(x) \, dx +
\inprod{u, \D p - p_s + p_t} 
-\frac{c}{2}\|\D p - p_s + p_t\|^2 \, ,
\eq
and derive the ALM-based continuous max-flow algorithm, see Alg.~\ref{seg-algd}.
An example of the foreground-background segmentation result of a 3D cardiac ultrasound image is shown by Fig.~\ref{fig:illus2} (b), which is computed by
the proposed continuous max-flow model \eqref{eq:seg-mf} based Alg.~\ref{seg-algd} (code is available at \cite{Cut-code}).  

\begin{algorithm}[!h]
\caption{ALM-Based Continuous Max-Flow Algorithm\label{seg-algd}} 
Initialize $u^0$ and $(p_s^0, p_t^0, p^0)$, for each iteration $k$ we explore the following two steps
\begin{itemize}
  \item fix $u^k$, compute $(p_s^{k+1},p_t^{k+1}, p^{k+1})$:
  \bq \label{alg:seg-stepp}
  (p_s^{k+1}, p_t^{k+1}, p^{k+1}) \, = \, \arg\max_{p_s,p_t, p} \, L_{c^k}(u^k, p_s, p_t, p) \, , \quad \text{s.t. \eqref{eq:seg-mf-fc1}}\, ,
  \eq
  provided the augmented Lagrangian function $L_c(p_s,p_t, p, u)$ in \eqref{eq:seg-auglag};
  \item fix $(p_s^{k+1},p_t^{k+1}, p^{k+1})$, then update $u^{k+1}$ by
  \bq 
  u^{k+1} \, :=\, u^k - c^k\big(\D p^{k+1} - p_s^{k+1} + p_t^{k+1}\big) \, . \label{alg:seg-stepu} 
  \eq
\end{itemize}
\end{algorithm}

\vspace{0.5cm}
\noindent{\bf Potts Model-Based Image Segmentation:}
For multiphase image segmentation, \emph{Pott model} is used as the basis to formulate the associate 
mathematical model~\cite{Boykov01anexperimental,Boykov01fastapproximate} by minimizing the following energy function
\bq \label{eq:potts-model} 
\min_{u} \, \sum_{i=1}^n
\int_{\Omega} u_i(x) \, \rho(l_i, x) \, dx \, + \, \alpha
\sum_{i=1}^n \int_{\Omega} \abs{\nabla u_i} \, dx \, 
\eq
subject to
\bq \label{eq:potts-bconst}
\sum_{i=1}^n \, u_i(x) \, = \, 1 \, , \quad { u_i(x) \in
\{0,1\}} \,, \; i = 1 \ldots n\,  , \quad \forall x \in \Omega \, ,
\eq
where $\rho(l_i, x)$, $i=1 \ldots n$, are the cost functions: for each pixel $x \in \Omega$, $\rho(l_i, x)$ gives
the cost to label $x$ as the segmentation region $i$. Potts model seeks the optimum labeling function $u_i^*(x)$, $i=1 \ldots n$, 
to the combinatorial optimization
problem~\eqref{eq:potts-model}, which defines the segmentation region $\Omega_i$ such that $u_i^*(x)=1$ for any $x \in \Omega_i$.
Clearly, the linear equality constraint $u_1(x) + \ldots + u_n(x) = 1$ states that each pixel $x$ belongs to a single segmentation region. 

Similar as the convex relaxed min-cut model \eqref{eq:seg-cr}, 
we can simply relax each binary constraint $u_i(x) \in \{0,1\}$ in \eqref{eq:potts-bconst} to 
the convex set $u_i(x) \in [0,1]$, then formulate the
convex relaxed optimization problem of Potts model \eqref{eq:potts-model} as
\bq \label{eq:relaxed-potts} 
\min_{u \in S} \, \sum_{i=1}^n
\int_{\Omega} u_i(x) \, \rho(l_i, x) \, dx \, + \, \alpha
\sum_{i=1}^n \int_{\Omega} \abs{\nabla u_i} \, dx \, 
\eq
where
\bq \label{eq:pw-simplex}
S \, = \, \{u(x) \,|\, (u_1(x) , \ldots, u_n(x)) \, \in \,
\triangle_n^{+} \, , \; \forall x \in \Omega \, \} \, .
\eq
$\triangle_n^{+}$ is the simplex set in the space $\R^n$. 

By simple variational analysis~\cite{YBTB10}, we have its equivalent primal-dual formulation
\begin{align} \label{eq:potts-pd}
\max_{p_s, p,q}\min_u\; & \underbrace{\int_{\Omega} p_s\, dx \, + \, \sum_{i=1}^n \int_{\Omega} u_i \, \big(p_i \, -
\, p_s   \, + \,  \D q_i \big) \, dx}_{L(u, p_s, p, q)} \\
\text{s.t. }\quad & \,  p_i(x) \, \leq \, \rho(\ell_i,x) \, , \quad
\abs{q_i(x)} \, \leq \, \alpha \, ; \quad i=1 \ldots n \,  \nonumber
\end{align}

Minimizing the energy function of \eqref{eq:potts-pd} over the free variable $u_i(x)$, it is easy to obtain its equivalent dual formulation, 
i.e. the \emph{continuous max-flow model}, for the convex relaxed Potts model~\eqref{eq:relaxed-potts} such that
\bq \label{eq:potts-mf}
\max_{p_s, p, q} \;  \int_{\Omega} p_s \,
dx  
\eq
subject to
\bq
\label{eq:mpf-cond-01} \abs{q_i(x)} \, \leq \, \alpha \, , \quad
p_i(x) \, \leq \, \rho(\ell_i,x) \, , \quad i=1 \ldots n \, ; \eq
\bq \label{eq:mpf-cond-03} \big( \D q_i - p_s + p_i \big)(x) \,
= \, 0 \, , \quad i=1,\ldots,n \, . 
\eq 

In view of the Lagrangian function $L(p_s, p, q, u)$ given in \eqref{eq:potts-pd}, we define its corresponding augmented Lagrangian function
\bq \label{eq:potts-auglag}
L_c(u, p_s, p, q) \, = \, \underbrace{\int_{\Omega} p_s\, dx \, + \, \sum_{i=1}^n \inprod{u_i, p_i  -
p_s  + \D q_i }}_{L(p_s, p, q, u)} \, - \, \frac{c}{2}\sum_{i=1}^n \norm{p_i  -
p_s  + \D q_i }^2 \, ,
\eq
and derive the ALM-based continuous max-flow algorithm, see Alg.~\ref{potts-algd}.
An example of the multiphase segmentation result of a 3D brain CT image is shown by Fig.~\ref{fig:illus2} (c), which is computed by
the proposed continuous max-flow model \eqref{eq:potts-mf} based Alg.~\ref{potts-algd} (code is available at \cite{Potts-code}).  

\begin{algorithm}[!h]
\caption{ALM-Based Continuous Max-Flow Algorithm\label{potts-algd}} 
Initialize $u^0$ and $(p_s^0, p_t^0, p^0)$, for each iteration $k$ we explore the following two steps
\begin{itemize}
  \item fix $u^k$, compute $(p_s^{k+1},p^{k+1}, q^{k+1})$:
  \bq \label{alg:potts-stepp}
  (p_s^{k+1}, p^{k+1}, q^{k+1}) \, = \, \arg\max_{p_s,p, p} \, L_{c^k}(u^k, p_s, p_t, q) \, ,
  \quad \text{s.t. \eqref{eq:mpf-cond-01}} \, ,
  \eq
  provided the augmented Lagrangian function $L_c(u,p_s,p, q)$ in \eqref{eq:potts-auglag};
  \item fix $(p_s^{k+1},p^{k+1}, q^{k+1})$, then update $u^{k+1}$ by
  \bq 
  u_i^{k+1} \, :=\, u_i^k - c^k\big(\D q_i^{k+1} - p_s^{k+1} + p_i^{k+1}\big)\, , \quad i \, = \, 1 \ldots n \, . \label{alg:potts-stepu} 
  \eq
\end{itemize}
\end{algorithm}

%% file: seg.tex
\section{Medical Image Segmentation}
\label{sec:seg}

Medical image segmentation is often much more challenging than segmenting 
camera photos, since medical imaging data usually suffers from low image quality, loss of imaging 
information, high
inhomogeneity of intensities and wrong imaging signals recorded etc. 
Prior knowledge about target regions is thus incorporated
into the related optimization models of medical image segmentation so as to improve the accuracy and robustness of
segmentation results and reduce manual efforts and intra- and inter-observer variabilities. 
In addition, the input 3D or 4D medical images often have a big data volume; therefore,
optimization algorithms with low iteration-complexity are appreciate in practice. With these respects, 
dual optimization approaches gained big successes in many applications of medical 
image segmentation~\cite{yuan2013efficient,qiu2014prostate,Ukwatta2015120, Ukwatta2012,ukwatta20133,rajchl2012fast,rajchl2014interactive,
Qiu2012,yuan2012efficient,Guo2015Globally,YUQRA2013,raey,Ukwatta2015120} etc.
In the following section, we will see a spectrum of priors can be easily integrated into
the introduced dual optimization framework without adding big efforts in numerics.

\subsection{Medical Image Segmentation with Volume-Preserving Prior}

\begin{figure}[!h]
\begin{center}
\begin{tabular}[p]{c@{\;}c@{\;}c}
\subfigure[]{
  \includegraphics[width=0.3\textwidth]{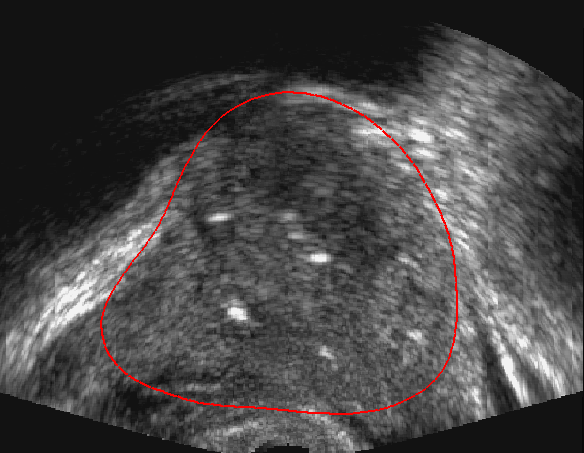}}&
  \subfigure[]{
  \includegraphics[width=0.3\textwidth]{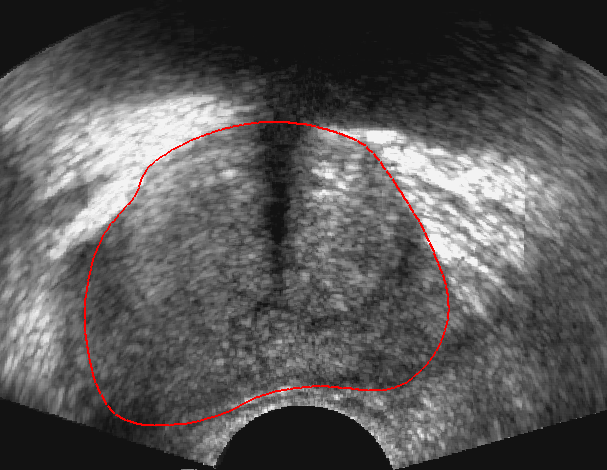}}&
  \subfigure[]{
  \includegraphics[width=0.265\textwidth]{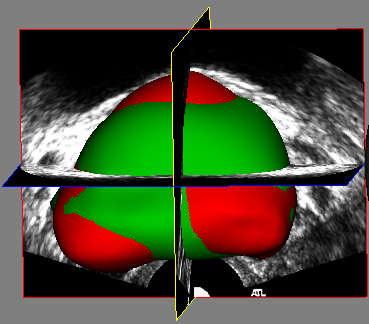}}
    \end{tabular}
 \end{center}
 \caption{\textit Example of 3D prostate TRUS segmentation: (a). sagittal view;
(b). coronal view; 
(c). segmentation result (green surface) overlapped with manual
segmentation (red surface).
 \label{fig:trus_vp}}
\end{figure}

Medical image data often has low image
quality, for example, the prostate transrectal ultrasound (TRUS) images (shown in Fig.~\ref{fig:trus_vp} (a) and (b)) usually with strong US speckles and shadowing
due to calcifications, missing edges or texture similarities between the inner and
outer regions of prostate. For segmenting such medical images, the volume information about the interesting
object region provides a global description for the image segmentation
task~\cite{raey}; on the other hand, such knowledge can be easily obtained in most cases
from learning the given training images or the other information sources.

To impose preserving the specified volum $\mathcal{V}$
in the continuous min-cut model \eqref{eq:seg-cr}, we penalize the difference between
the volume of the target segmentation region and $\mathcal{V}$ such that
\bq \label{eq:seg-vp}
\min_{u(x) \in [0,1]} \; \int_{\Omega} \big\{ \big(1-u \big)C_s  + u \, C_t \big\} \, dx \, +\,
\alpha \int_{\Omega}\abs{\nabla u} \, dx \, +  \, \gamma \abs{\mathcal{V} - \int_{\Omega} u \, dx}\, .
\eq

By means of the conjugate expression of the absolute function such that 
$\gamma \abs{v} = \max_{r \in [-\gamma, \gamma]} r \cdot v$, we have
\[
\gamma \abs{\mathcal{V} - \int_{\Omega} u \, dx} \, = \, \max_{r \, \in \, [-\gamma, \gamma]} \,
r \big(\mathcal{V} - \int_{\Omega} u \, dx \big) \, .
\]

With variational analysis, it can be provded that the volume-preserving min-cut model \eqref{eq:seg-vp} can
equally represented by~\cite{raey} 
\begin{align}
\max_{p_s, p_t, p, \lambda} \; & \int_{\Omega} p_s(x) \, dx  + {r \mathcal{V}}\,  \label{eq:mf-vp}\\
\textbf{s.t.} \; \;& \abs{p(x)}   \leq \alpha \, ,  \quad 
 p_s(x) \leq  C_s(x) \, , \quad
p_t(x) \leq  C_t(x) \,  , \quad r \, \in \, [-\gamma, \gamma] \, ; \label{eq:mf-vp-const1} \\
& \Big(\D p -p_s + p_t \Big)(x) - r
\, = \, 0 \label{eq:mf-vp-const}\, .
\end{align}

In contrast to the linear equality constraint \eqref{eq:seg-mf-fc2}, i.e. the exact flow balance constraint in the maximal flow
setting \eqref{eq:seg-mf}, such exact flow balance constraint is relaxed to be within a range of $r \in [-\gamma, \gamma]$ as shown in \eqref{eq:mf-vp-const}. 
In addition, the value $r$ is penalized in maximum flow configuration as shown in \eqref{eq:mf-vp}.

Under such dual optimization perspective, the optimum labelling $u^*(x)$ to \eqref{eq:seg-vp} works just as the optimal multiplier to the linear equality
constraint \eqref{eq:mf-vp-const}, i.e. $\Big(\D p -p_s + p_t \Big)(x) - r = 0$ and $r \in [-\gamma, \gamma]$. 

The same as \eqref{eq:seg-auglag}, we define the augmented Lagrangian function w.r.t. \eqref{eq:mf-vp}
\bq \label{eq:vp-auglag}
L_c(u, p_s,p_t, p, r) \, = \,\int_{\Omega} \, p_s(x) \, dx +
\inprod{u, \D p - p_s + p_t - r} 
-\frac{c}{2}\|\D p - p_s + p_t - r\|^2 \, ,
\eq
and derive the related ALM-based algorithm, see Alg.~\ref{vp-algd}.

Example of segmenting 3D prostate TRUS images demonstrated, as shown in Fig.~\ref{fig:trus_vp}, the segmentation
results with the volume-preserving prior \eqref{eq:seg-vp} significantly improves the results without the volume prior from DSC $78.3 \pm 7.4 \%$ 
to $89.5 \pm 2.4 \%$ in DSC~\cite{raey}.

\begin{algorithm}[!h]
\caption{ALM-Based Segmentation Algorithm with Volume-Preserving Prior \label{vp-algd}} 
Initialize $u^0$ and $(p_s^0, p_t^0, p^0, r^0)$, for each iteration $k$ we explore the following two steps
\begin{itemize}
  \item fix $u^k$, compute $(p_s^{k+1},p_t^{k+1}, p^{k+1}, r^{k+1})$:
  \bq \label{alg:vp-stepp}
  (p_s^{k+1}, p_t^{k+1}, p^{k+1}, r^{k+1}) \, = \, \arg\max_{p_s,p_t, p, r} \, L_{c^k}(u^k, p_s, p_t, p, r) \, ,
  \quad \text{s.t.  \eqref{eq:mf-vp-const1}}
  \eq
  provided the augmented Lagrangian function $L_c(u, p_s,p_t, p, r)$ in \eqref{eq:vp-auglag};
  \item fix $(p_s^{k+1},p_t^{k+1}, p^{k+1}, r^{k+1})$, then update $u^{k+1}$ by
  \bq 
  u^{k+1} \, :=\, u^k - c^k\big(\D p^{k+1} - p_s^{k+1} + p_t^{k+1} - r^{k+1} \big) \, . \label{alg:vp-stepu} 
  \eq
\end{itemize}
\end{algorithm}

\subsection{Medical Image Segmentation with Compactness Priors}

\begin{figure}[!h]
\begin{center}
\begin{tabular}[p]{c@{\;}c@{\;}c}
\subfigure[]{
  \includegraphics[width=0.25\textwidth]{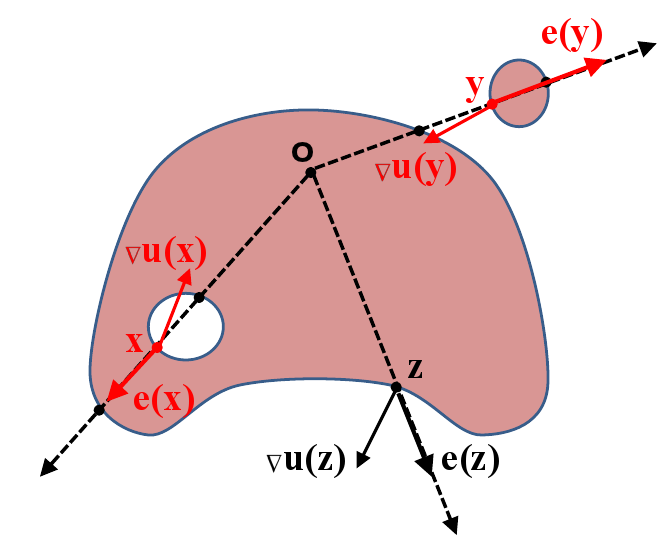}}&
  \subfigure[]{
  \includegraphics[width=0.25\textwidth]{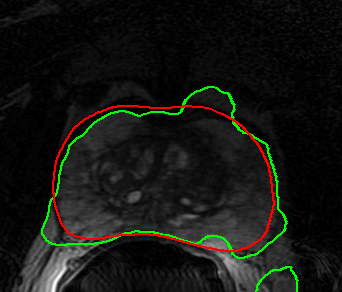}}&
  \subfigure[]{
  \includegraphics[width=0.25\textwidth]{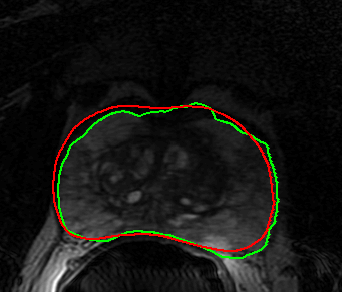}}
    \end{tabular}
 \end{center}
 \caption{\textit (a). illustration of compactness (star-shape) prior;
(b). the segmentation result of a 3D prostate MRI with compactness prior; 
(c). the segmentation result of a 3D prostate MRI without compactness prior.
 \label{fig:star}}
\end{figure}

The star-shape prior is a powerful description of region shapes, which enforces the segmentation region to be
compact, i.e. a single region without any cavity, namely the compactness prior.
Usually, the star-shape prior is defined with respect to a center point $O$ (see Fig.~\ref{fig:star} (b)): an object has a
star-shape if for any pixel $x$ inside the object, all points on the straight line
between the center $O$ and $x$ also lie inside the object; in another word, the object
boundary can only pass any radial line starting from the origin $O$ one single
time. 

To formulate such a compactness prior, let $d_O(x)$ be the
distance map with respect to the origin point $O$ and $e(x) = \nabla d_O(x)$. Then the compactness prior can be
defined as
\bq \label{eq:stars}
(\nabla u \, \cdot \,  e)(x) \, \geq \, 0 \, .
\eq

Now we integrate the compactness prior \eqref{eq:stars} into the continuous min-cut model \eqref{eq:seg-cr}, which results in
the following image segmentation model:
\bq \label{eq:seg-star}
\min_{u(x) \in [0,1]} \; \int_{\Omega} \big\{ \big(1-u \big)C_s  + u \, C_t \big\} \, dx \, + \,
\alpha \int_{\Omega}\abs{\nabla u} \, dx \, , \quad \textbf{s.t.}\; \; (\nabla u \cdot  e)(x) \, \geq \, 0 \, ,
\eq
which is indentical to
\bq \label{eq:seg-star-1}
\max_{\lambda(x)\leq 0} \min_{u(x) \in [0,1]} \; \int_{\Omega} \big\{ \big(1-u \big)C_s  + u \, C_t \big\} \, dx\, + \,
\alpha \int_{\Omega}\abs{\nabla u} \, dx \, + \, \int_{\Omega} \lambda(x) (\nabla u \cdot  e)(x)\, dx \, .
\eq

Yuan et al \cite{yuan2012efficient} showed that, with similar variational analysis as in~\cite{Yuan2010,ybtb2014}, the convex optimization
model \eqref{eq:seg-star} is equivalent to the dual formulation below
\begin{align}
\max_{p_s, p_t, p, \lambda} \; & \int_{\Omega} p_s(x) \, dx \,   \label{eq:dual-mf-star}\\
\textbf{s.t.} \;\;& \abs{p(x)}   \leq \alpha \, ,  \quad 
 p_s(x) \leq  C_s(x) \, , \quad
p_t(x) \leq  C_t(x) \,  ; \label{eq:dual-star-const2} \\
& \Big(\D \big(p - { \lambda e} \big) -p_s + p_t \Big)(x)
\, = \, 0 \, , \quad { \lambda(x) \leq 0} \,  . \label{eq:dual-star-const}
\end{align}
With this perspective, the optimum labelling function $u^*(x)$ works just as the optimal multiplier to the linear equality
constraint \eqref{eq:dual-star-const}.

Clearly, the dual optimization model \eqref{eq:dual-mf-star} is similar as the continuous max-flow model \eqref{eq:seg-mf}, with just an additional
flow variable $\lambda(x)$ subject to the constraint $\lambda(x) \leq 0$. 
The same as \eqref{eq:seg-auglag}, we define the augmented Lagrangian function w.r.t. \eqref{eq:dual-mf-star}
\bq \label{eq:star-auglag}
L_c(u, p_s,p_t, p, \lambda) \, = \,\int_{\Omega} \, p_s(x) \, dx +
\inprod{u, \D \big(p - { \lambda e} \big) - p_s + p_t} 
-\frac{c}{2}\|\D \big(p - { \lambda e} \big) - p_s + p_t\|^2 \, ,
\eq
and derive the related ALM-based algorithm, see Alg.~\ref{star-algd}.

An example of the segmentation of a 3D prostate MR image is shown by Fig.~\ref{fig:star} (b) and (c), with and without 
the compactness prior respectively. It is easy to see that some segmentation bias region is introduced in the result, which is in contrast 
to the segmentation result with the compactness prior.

\begin{algorithm}[!h]
\caption{ALM-Based Segmentation Algorithm with Compactness Prior \label{star-algd}} 
Initialize $u^0$ and $(p_s^0, p_t^0, p^0, \lambda^0)$, for each iteration $k$ we explore the following two steps
\begin{itemize}
  \item fix $u^k$, compute $(p_s^{k+1},p_t^{k+1}, p^{k+1}, \lambda^{k+1})$:
  \bq \label{alg:star-stepp}
  (p_s^{k+1}, p_t^{k+1}, p^{k+1}, \lambda^{k+1}) \, = \, \arg\max_{p_s,p_t, p, \lambda} \, L_{c^k}(u^k, p_s, p_t, p, \lambda) \, ,
  \quad \text{s.t.  \eqref{eq:dual-star-const2}}
  \eq
  provided the augmented Lagrangian function $L_c(u,p_s,p_t, p, \lambda)$ in \eqref{eq:star-auglag};
  \item fix $(p_s^{k+1},p_t^{k+1}, p^{k+1}, \lambda^{k+1})$, then update $u^{k+1}$ by
  \bq 
  u^{k+1} \, :=\, u^k - c^k\big(\D \big(p^{k+1} - { \lambda^{k+1} e} \big) - p_s^{k+1} + p_t^{k+1}\big) \, . \label{alg:star-stepu} 
  \eq
\end{itemize}
\end{algorithm}

\subsection{Medical Image Segmentation with Region-Order Prior}

In many practices of medical image segmentation, the target regions have exact inter-region relationships in geometry, for example, 
one region is contained in another region as its subregion. Such inclusion/overlay order between regions, a.k.a. region-order, appears quite often in medical image segmentation, such as
the three regions of blood pool, myocardium and background are overlaid sequentially in 3D cardiac T2 MRI~\cite{rajchl2012fast,Nambakhsh20131010}, the central zone
is well included inside the whole gland region of prostate in 3D T2w prostate MRIs~\cite{YUQRA2013},
the region inside carotid artery adventitia boundary(AB) covers the region inside lumen-intima
boundary(LIB) in input T1-weighted
black-blood carotid magnetic resonance (MR)
images\cite{Ukwatta2012,ukwatta20133} etc. 
In practice, imposing such geometrical order for exacting target regions can significantly improve both accuracy and robustness of image segmentation. 

\begin{figure}[!h]
\begin{center}
\begin{tabular}[p]{c@{\,}c@{\,}c@{\,}c}
\subfigure[]{
  \includegraphics[height=2.6cm]{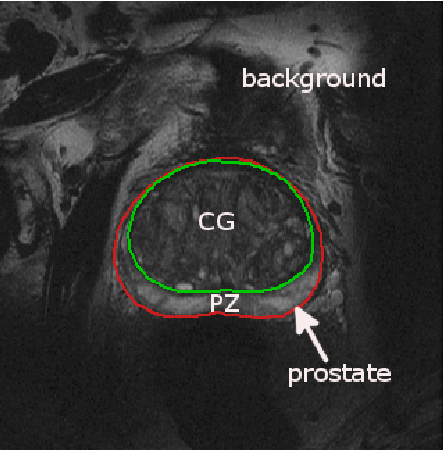}}&
  \subfigure[]{
  \includegraphics[height=2.6cm]{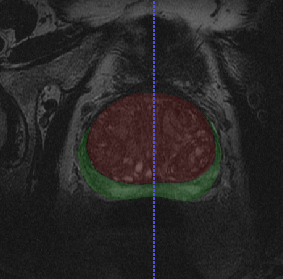}
  \includegraphics[height=2.6cm]{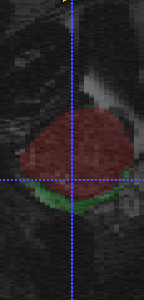}
  }&
  \subfigure[]{
  \includegraphics[height=2.6cm]{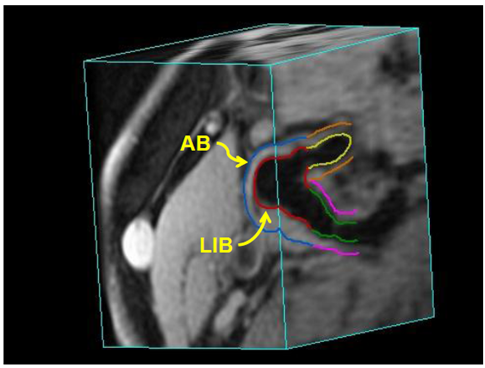}}&
  \subfigure[]{
  \includegraphics[height=2.6cm]{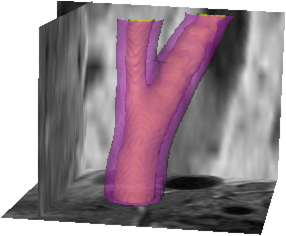}}
    \end{tabular}
 \end{center}
 \caption{\textit (a). illustration of the segmented contours overlaid on a T2w prostate MRI slice, where the central zone (CZ) of prostate (inside the green contour)
 is included in the whole prostate region (inside the red contour); (b). the segmentation result in axial and saggital views respectively~\cite{YUQRA2013};
(c). illustration of the overlaid surfaces by adventitia (AB) and lumen-intima (LIB) in 3D carotid MR images; 
(d). the result of segmented surfaces of AB and LIB in the input 3D carotid MR image \cite{Ukwatta2012,ukwatta20133}.
 \label{fig:lo}}
\end{figure}

Such overlaid regions, also linear-ordered regions, can be mathematically formulated as  
\bq \label{eq:linear-ro}
\Omega_n (\:=\emptyset) \, \subseteq \, \Omega_{n-1} \, \subseteq \, \ldots \, \subseteq \, \Omega_{1} \, \subseteq \, \Omega_0 (:=\Omega)  \, .
\eq

In view of Potts model-based multiphase image segmentation model~\eqref{eq:potts-model}, we can therefore encode the total segmentation 
cost and surface regularization terms as the following \emph{coupled continuous min-cut model}:
\begin{align}
 \min_{u_i(x) \in [0,1]}\; & \; \sum\limits_{i=1}^n \int_\Omega (u_{i-1} - u_i) D_i\, dx + 
			   \sum\limits_{i=1}^{n-1} \alpha_i \int_\Omega  | \nabla u_i |\, dx  \, ,  \label{eq:coupled-mc}\\
 \text{s.t.} \; & \; 
			  0 \, \leq \, u_{n-1}(x)  \, \leq \, \ldots \, \leq u_1(x) \, \leq \, 1 \, ; \label{eq:cmc-const} 
\end{align}
where $u_i(x)$, $i=0 \ldots n$, is the indicator function of the respective region $\Omega_i$ and 
$D_i(x)$ is the cost for the pixel $x$ inside the region $\Omega_{i-1} \backslash \Omega_i$ which is labelled by $u_{i-1}(x)-u_i(x)$.

It can be proved that, with simple variational computation, the coupled continuous min-cut model~\eqref{eq:coupled-mc}
can be equivalently reformulated as the followed dual optimization problem~\cite{Bae2010b}
\begin{align}
\max_{p_i, q_i} \; & \;  \int_{\Omega} p_1 \, dx \label{eq:cmf} \\
\text{s.t.} \;& \; \abs{q_i(x)} \, \leq \, \alpha_i
\, , \; p_i(x) \, \leq \, D_i(x) \, , \label{eq:cmf-const1} \\
& \big( \D q_i - p_{i} + p_{i+1}\big)(x)
\, = \, 0 \, , \; i=1 \ldots n-1 \,  .  \label{eq:cmf-const2}
\end{align}

Also, for the dual optimization model~\eqref{eq:cmf}, the optimum labelling function $u_i^*(x)$, $i=1 \ldots n-1$, 
works as the optimal multiplier to the respective linear equality
constraint \eqref{eq:cmf-const2}, which can be seen from the corresponding Lagrangian function of \eqref{eq:cmf}:
\bq \label{eq:cmf-lag}
L(u, p, q) \, = \, \int_{\Omega} p_1 \, dx \, + \, \sum_{i=1}^{n-1} \inprod{u_i, \D q_i - p_{i} + p_{i+1}} \, .
\eq

Similarly, we define the augmented Lagrangian function w.r.t. \eqref{eq:cmf-lag}
\bq \label{eq:cmf-auglag}
L_c(u, p, q) \, = \,\int_{\Omega} \, p_s(x) \, dx  \, + \, \sum_{i=1}^{n-1} \inprod{u_i, \D q_i - p_{i} + p_{i+1}}\,
- \, \frac{c}{2} \sum_{i=1}^{n-1} \norm{ \D q_i - p_{i} + p_{i+1}}^2 \, ,
\eq
and derive its related ALM-based algorithm, see Alg.~\ref{ol-algd}.

Two examples are given in Fig.~\ref{fig:lo}, (b). demonstrate that the two regions of the prostate central zone (CZ) and whole gland (WG)
are extracted the segmentation from the given 3D T2w prostate MR image~\cite{YUQRA2013} subject to the enforced linear
region-order constraint $\Omega_{PZ} \subset \Omega_{WG}$; (d). show that the two regions of carotid lumen, i.e.
contoured by AB and LIB, are well segmented by imposing the linear region-order constraint $\Omega_{LIB} \subset \Omega_{AB}$~\cite{Ukwatta2012,ukwatta20133}.

\begin{algorithm}[!h]
\caption{ALM-Based Segmentation Algorithm with Linear Region-Order Prior \label{ol-algd}} 
Initialize $u^0$ and $(p^0, q^0)$, for each iteration $k$ we explore the following two steps
\begin{itemize}
  \item fix $u^k$, compute $(p^{k+1},q^{k+1})$:
  \bq \label{alg:ol-stepp}
  (p^{k+1}, q^{k+1}) \, = \, \arg\max_{p,q} \, L_{c^k}(u^k, p, q) \, , \quad \text{s.t.  \eqref{eq:cmf-const1}}
  \eq
  provided the augmented Lagrangian function $L_c(u,p,q)$ in \eqref{eq:cmf-auglag};
  \item fix $(p^{k+1},q^{k+1})$, then update $u^{k+1}$ by
  \bq 
  u_i^{k+1} \, :=\, u_i^k - c^k\big(\D q_i^{k+1} - p_{i}^{k+1} + p_{i+1}^{k+1} \big) \, . \label{alg:ol-stepu} 
  \eq
\end{itemize}
\end{algorithm}

\subsection{Extension to Partially Ordered Regions}

\begin{figure}[!h]
\begin{center}
\begin{tabular}[p]{c@{\,}c}
  \subfigure[]{
  \includegraphics[height=2.75cm]{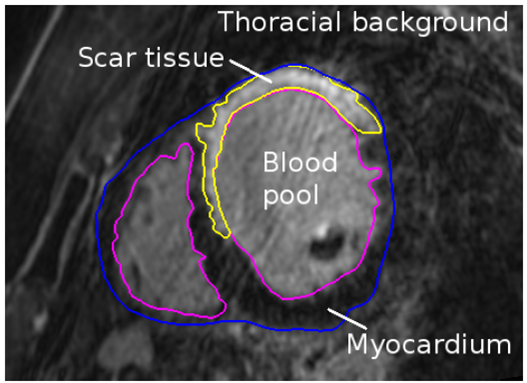}
  \includegraphics[height=2.75cm]{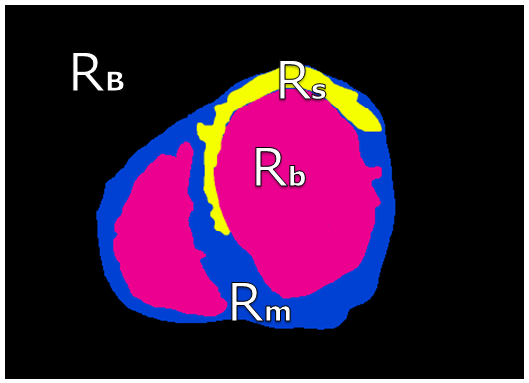}
  }&
  \subfigure[]{
  \includegraphics[height=2.75cm]{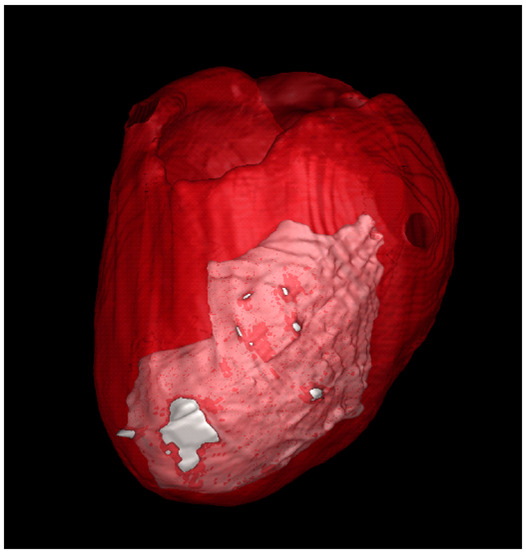}
  \includegraphics[height=2.75cm]{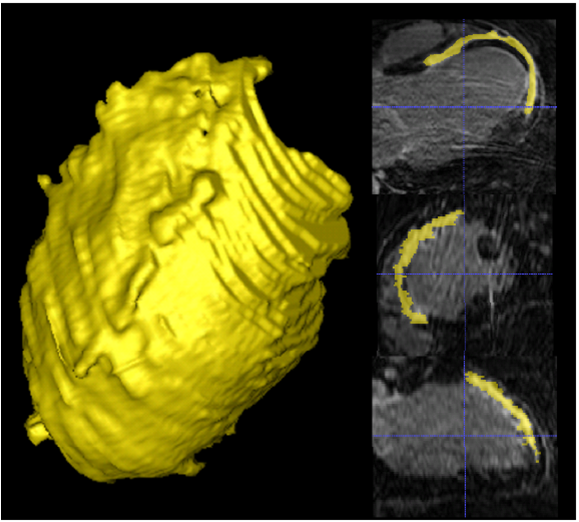}
  }
    \end{tabular}
 \end{center}
 \caption{\textit (a). Illustration of the anatomical spatial order of cardiac regions in a
 LE-MRI slice: the region $R_C$ containing the heart is divided into three sub-regions including
 myocardium $R_m$, blood $R_b$, and scar tissue $R_s$. $R_B$ represents the thoracical background
 region. (b). 3D segmentation results of LE-MRI: myocardium $R_m$ (red) and scar tissue $R_s$ (yellow).
 \label{fig:po}}
\end{figure}

An extension to the linear-order of regions~\eqref{eq:linear-ro} is the partial-order of regions, for which the geometric inter-region relationship
can be often formulated as
\[
\Omega_{k} \, \supset \, \Omega_{1} \, \cup \, \ldots \, \cup \, \Omega_{k-1}\, , \quad \text{or }\quad 
\Omega_{k} \, = \, \Omega_{1} \, \cup \, \ldots \, \cup \, \Omega_{k-1} \, .
\]

For example, the segmentation of 3D LE-MRIs~\cite{Rajchl2012A,rajchl2014interactive} targets to extract the thoracical background
region $R_B$ and its complementary region of the whole heart $R_C$ in the input
LE-MRIs
\[
\Omega \, = \, R_C \, \cup \, R_B \, , \quad R_C \, \cap \, R_B \, = \, \emptyset \, ,
\] 
and the cardiac region $R_c$ contains three sub-regions of
myocardium $R_m$, blood $R_b$, and scar tissue $R_s$ (see Fig. \ref{fig:po} (a) for illustrtaion):
\bq
\label{eq:scar-subregions}  
R_C \, = \, \Big( R_{m} \, \cup \, R_{b} \, \cup \,
R_{s} \Big) \, 
\eq 
where the three sub-regions $R_m$, $R_b$ and $R_s$ are mutually disjoint 
\bq \label{eq:scar-disjr}
R_m \cap R_b \, = \, \emptyset \, , \quad R_b \cap R_s \, = \, \emptyset \,
, \quad R_s \cap R_m \, = \, \emptyset \, .
\eq

The same as Potts model \eqref{eq:relaxed-potts}, let $u_i(x) \in \{0,1\}$, $i \in m,b,s,C,B\}$, be the indicator function of the region $R_i$,
then the region-order constraints \eqref{eq:scar-subregions} and \eqref{eq:scar-disjr} can be expressed as
\bq \label{eq:po-rsum}
u_C(x) \, + \, u_B(x) \, = \, 1\, , \quad u_C(x) \, = \, u_m(x) \, + \, u_b(x) \, + \, u_s(x) \, , \quad \forall x \, \in \, \Omega \, ,
\eq
the associate image segmentation model is formulated as  
\bq \label{eq:po-potts} 
\min_{u(x) \in \{0,1\}} \, \sum_{i \in \{m,b,s,B\} }\, 
\int_{\Omega} u_i(x) \, \rho(l_i, x) \, dx \, + \, \alpha \,
\sum_{i \in \{m,b,s,B,C\}} \int_{\Omega} \abs{\nabla u_i} \, dx \, 
\eq
where the first term sums up the costs of four disjoint segmentation regions $R_{m,.b,s,B}$, and the second
term regularizes the surfaces of all the regions including $R_{m,b,s,B,C}$.

Through variational computation~\cite{Rajchl2012A,rajchl2014interactive}, we obtain the equivalent dual model to 
the convex relaxation of \eqref{eq:po-potts}
\bq \label{eq:po-potts-mf}
\max_{p_o, p, q} \;  \int_{\Omega} p_o \, dx  
\eq
subject to
\bq
\label{eq:po-mpf-cond01} \abs{q_i(x)} \, \leq \, \alpha \, , \quad i \in \{m,b,s,B,C\}\, ; \quad 
p_i(x) \, \leq \, \rho(\ell_i,x) \, , \quad i \in \{ m,b,s,B\} \, ; \eq
\bq \label{eq:po-mpf-cond03} \big( \D q_i - p_o + p_i \big)(x) \,
= \, 0 \, , \quad i \in \{B, C\} \, ; 
\eq 
\bq \label{eq:po-mpf-cond02} \big( \D q_i - p_C + p_i \big)(x) \,
= \, 0 \, , \quad i \in \{m,b,s\} \, .
\eq 

The labelling function $u_i(x)$, $i \in \{m,b,s,B,C\}$, works as the multiplier to the linear
equality constraint \eqref{eq:po-mpf-cond03} - \eqref{eq:po-mpf-cond02} respectively.
This hence gives the clue to define the related augmented Lagrangian function and 
build up the similar ALM-based optimization algorithm as Alg.~\ref{potts-algd} which is omitted here (see~\cite{Rajchl2012A,rajchl2014interactive} for more details).

Actually, more complex priors of region-orders can be defined and employed in medical image segmentation, see \cite{Baxter2017Directed} for references.

\subsection{Medical Image Segmentation with Spatial Consistency Prior}

For another kind of medical image segmentation tasks, the target regions usually appear with spatial similarities between two neighbour slices or multiple co-registered
volumes, for example, 3D prostate ultrasound image
segmentation \cite{yuan2013efficient,qiu2014prostate,Ukwatta2015120},
co-segmenting lung pulmonary $^1$H and hyperpolarized $^3$He MRIs
\cite{Guo2015Globally}. This greatly helps making full use of image features in all related images and guides the simultaneous segmentation procedure to
reach a higher accuracy in result.

\begin{figure}[!h]
\begin{center}
\begin{tabular}[p]{c@{\hspace{0.5cm}}c}
\subfigure[]{
  \includegraphics[height=4.68cm]{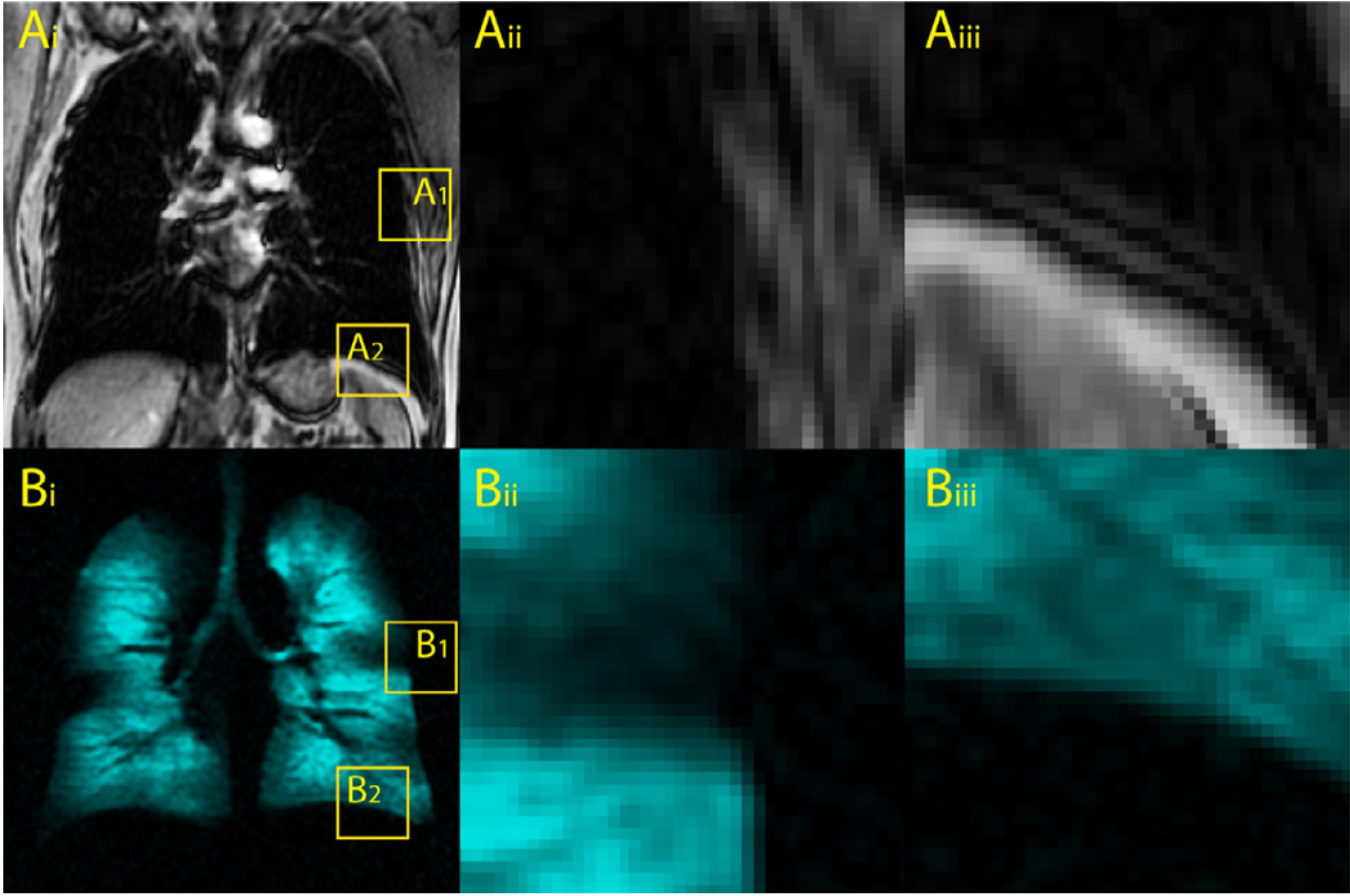}}&
  \subfigure[]{
  \includegraphics[height=4.68cm]{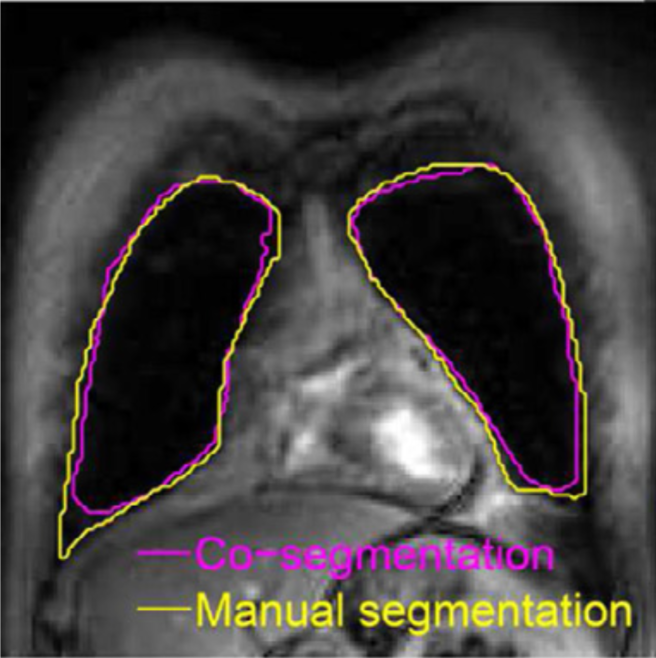}}
    \end{tabular}
 \end{center}
 \caption{\textit (a). Complementary edge information about $^1$H and $^3$He 3D lung MRIs; 
 (Ai) $^1$H MRI coronal slice with inset boxes: A1 expanded in (Aii ) and A2 expanded in (Aiii ); 
 (Bi ) $^3$He MRI coronal slice with inset boxes: B1 expanded in (Bii) and B2 expanded in (Biii).
 (b). The co-segmentation result shown in a $^1$H MRI coronal slice (pink contour) w.r.t. the manual segmentation
 result (yellow contour), see \cite{Guo2015Globally} for more details.
 \label{fig:coseg}}
\end{figure}

Now we consider jointly segmenting two input images \cite{Guo2015Globally}, i.e. co-segmentation, for simplicities. 
Given two appropriately co-registered images (see Fig.~\ref{fig:coseg} (a)), we propose to simultaneously
extract the same target region from both images, i.e., $\R^1$ and $\R^2$, and let $\R_B^1$ and $\R_B^2$ be the associate
complementary background regions. Initially, the labelling function $u_i(x)$ for each 
target region $\R^i$, $i=1,2$, can be computed through the continuous min-cut model
\eqref{eq:seg-cr}, i.e.
\bq \label{eq:seg-cr}
\min_{u_i(x) \in [0,1]} \;
 \int_{\Omega} \big\{ \big(1-u_i \big)C_s^i  + u_i \, C_t^i \big\} \, dx \, + \, \alpha \int_{\Omega}\abs{\nabla u_i}\, dx \, ,
\eq
In addition, we impose the spatial similarity between the two target regions
$\R^1$ and $\R^2$ by penalizing the total difference of
$\R^1$ and $\R^2$, i.e. $\int_{\Omega} \abs{u_1 - u_2}\, dx$. Hence, we have
the convex optimization model for co-segmenting the two input images:
\bq \label{eq:coseg-cr}
\min_{u_{1,2}(x) \in [0,1]} \; 
\sum_{i=1}^2 \int_{\Omega} \big\{ \big(1-u_i \big)C_s^i  + u_i \, C_t^i \big\} \, dx \, + \, \alpha \sum_{i=1}^2 \int_{\Omega}\abs{\nabla u_i}\, dx
\, + \, \beta \int_{\Omega} \abs{u_1 - u_2} \, dx \, . 
\eq

Observe that
\[
\beta \int_{\Omega} \abs{u_1 - u_2}\, dx \, = \, \max_{r(x) \in [- \beta, \beta]} \, \int_{\Omega} r \, (u_1 - u_2)\, dx \, ,
\]
and similar variational analysis as in~\cite{Yuan2010,ybtb2014}, we have the equivalent representation of the co-segmentation optimization model~\eqref{eq:coseg-cr}
\begin{align}
\min_{u_{1,2}} \max_{p_s, p_t, q, r} \; L(u, p_s, p_t, q, r) \, = \,  \nonumber &
\int_{\Omega} p_s^1 \, dx \, + \,
\int_{\Omega} p_s^2 \, dx \, + \,
\inprod{u_1, \D q_1  + p_t^1  -  p_s^1
+ r } \, +  \, \nonumber \\
& \inprod{u_2, \D q_2 + p_t^2 -  p_s^2 - r} \, . \label{eq:coseg-pd}
\end{align}

Minimizing the primal-dual optimization model \eqref{eq:coseg-pd} over $u_1(x)$ and $u_2(x)$ first, we then derive
the dual optimization model to~\eqref{eq:coseg-cr} such that
\bq \label{eq:coseg-mf}
\max_{p_s^{1,2}, p_t^{1,2}, q_{1,2}, r} \; \int_{\Omega} p_s^1 \, dx \, + \,
\int_{\Omega} p_s^2 \, dx
\eq
subject to
\bq \label{eq:coseg-mf-fc}
p_s^i(x) \, \leq \, C_s^i(x) \, , \quad 
p_t^i(x) \, \leq \, C_t^1(x) \, , \quad
\abs{q_i(x)} \, \leq \, \alpha \, , \quad i \, = 1,2 \, ; \quad
\abs{r(x)} \, \leq \, \beta \, ;
\eq
\bq \label{eq:coseg-mf-const}
\D q_1(x) \, + \, p_t^1(x) \, - \, p_s^1(x) \,
+ \, r(x) \, = \, 0 \, , \quad
\D q_2(x) \, + \, p_t^2(x) \, - \, p_s^2(x) \,
- \, r(x) \, = \, 0 \, .
\eq
The optimum $u_1^*(x)$ and $u_2^*(x)$ to the convex optimization model \eqref{eq:coseg-cr} are exactly the optimum multipliers to
the linear equality constraints \eqref{eq:coseg-mf-const} respectively.

Correspondingly, we define the augmented Lagrangian function w.r.t. $ L(u, p_s, p_t, q, r)$ in \eqref{eq:coseg-pd}
\bq \label{eq:coseg-auglag}
L_c(u, p_s,p_t, q, r) \, = \, L(u, p_s, p_t, q, r)
\, -\, \frac{c}{2} \|\D q_1 - p_s^1 + p_t^1  + r \|^2 \, 
- \, \frac{c}{2} \|\D q_2 - p_s^2 + p_t^2  - r \|^2 \, .
\eq
and derive the ALM-based image co-segmentation algorithm, see Alg.~\ref{coseg-algd}.

\begin{algorithm}[!h]
\caption{ALM-Based Image Co-Segmentation Algorithm\label{coseg-algd}} 
Initialize $u^0$ and $(p_s^0, p_t^0, q^0, r^0)$, for each iteration $k$ we explore the following two steps
\begin{itemize}
  \item fix $u^k$, compute $(p_s^{k+1},p_t^{k+1}, q^{k+1}, r^{k+1})$:
  \bq \label{alg:coseg-stepp}
  (p_s^{k+1}, p_t^{k+1}, q^{k+1}, r^{k+1}) \, = \, \arg\max_{p_s,p_t, q, r} \, L_{c^k}(u^k, p_s, p_t, q, r) \, ,
  \quad \text{s.t.  \eqref{eq:coseg-mf-fc}} \, 
  \eq
  provided the augmented Lagrangian function $L_c(u, p_s,p_t, q, r)$ in \eqref{eq:coseg-auglag};
  \item fix $(p_s^{k+1},p_t^{k+1}, q^{k+1}, r^{k+1})$, then update $u^{k+1}$ by
  \begin{align}
  u_1^{k+1} \, :=& \, u_1^k - c^k\big(\D q_1^{k+1} - (p_s^1)^{k+1} + (p_t^1)^{k+1} + r \big)\,  , \nonumber \\
    u_2^{k+1} \, := &\, u_2^k - c^k\big(\D q_2^{k+1} - (p_s^2)^{k+1} + (p_t^2)^{k+1} - r \big) \, . \nonumber
  \end{align}
\end{itemize}
\end{algorithm}

Clearly, without $r(x)$, the dual model \eqref{eq:coseg-mf} of image co-segmentation can be viewed as two independent continuous max-flow models \eqref{eq:seg-mf}; and
the associate ALM-based image co-segmentation algorithm (Alg.~\ref{coseg-algd}) is just two separate continuous max-flow algorithms (Alg.~\ref{seg-algd}) in combination.
Observing this, we see that the ALM-based image co-segmentation algorithm (Alg.~\ref{coseg-algd}) is exactly two joint
continuous max-flow algorithms (Alg.~\ref{seg-algd}) along with optimizing the additional variable $r(x)$. This is much simpler than solving
the convex optimization model \eqref{eq:coseg-cr} directly!

An example of image co-segmentation of $^1$H and $^3$He 3D lung MRIs is shown in Fig~\ref{fig:coseg}. The results showed that 
the introduced image co-segmentation approach \eqref{eq:coseg-cr} and \eqref{eq:coseg-mf} yields 
superior performance compared to the single-channel image segmentation in terms of precision, accuracy and robustness \cite{Guo2015Globally}.

The introduced optimization method can be easily extended to simultaneously segmenting a series of images while enforcing
their spatial consistencies between images~\cite{yuan2013efficient,qiu2014prostate,Ukwatta2015120}. For example, it can be used to enforce the axial symmetry prior between 3D prostate TRUS image slices~\cite{yuan2013efficient,qiu2014prostate},
the prior can be formulated as
\[
 \sum_{i=1}^{n-1} \int_{\Omega} \abs{u_{i+1} -
u_i}\, dx \, + \, \int_{\Omega} \abs{u_{n}(L-x) - u_1(x)} dx  \, ,
\]
i.e. a sequence of spatial consistencies between the given image slices to be enforced. Similarly, it results in
a series of joint continuous max-flow computations (see~\cite{yuan2013efficient,qiu2014prostate}).

\begin{figure}[!h]
\begin{center}
\begin{tabular}[p]{c@{\hspace{0.5cm}}c}
\subfigure[]{
  \includegraphics[height=3.9cm]{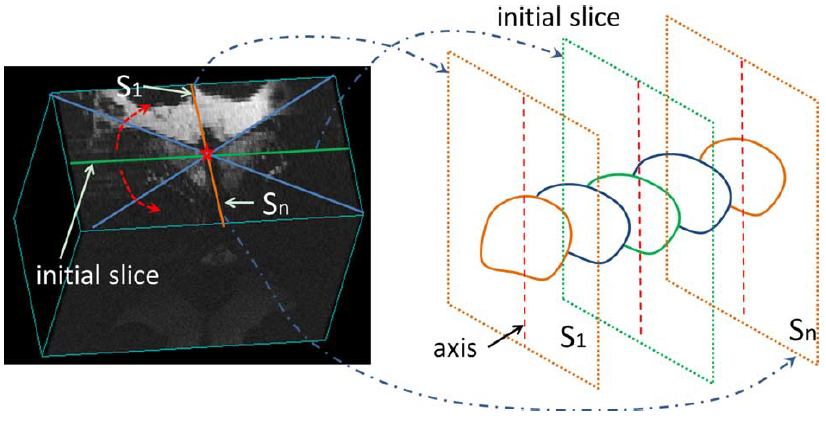}}&
  \subfigure[]{
  \includegraphics[height=3.9cm]{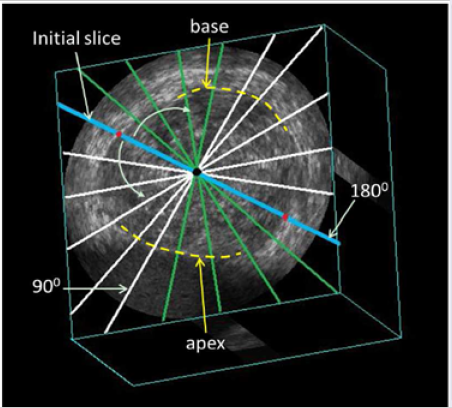}}
    \end{tabular}
 \end{center}
 \caption{\textit (a). illustration of 3D prostate TRUS image slices; 
 (b). the joint segmentation of all slices with the axial symmetry prior, see \cite{yuan2013efficient,qiu2014prostate} for more details.
 \label{fig:coseg-2}}
\end{figure}

%% file: reg.tex
\section{Non-rigid Medical Image Registration}
\label{sec:reg}

\subsection{Sequential Convex Optimization for Medical Image Registration}
In this section, we introduce the intensity-based non-rigid medical image registration
model, which is based the well-known optical flow model~\cite{Horn81determiningoptical,Barron1994}
to align the given image
pair $I_f(x)$ and $I_r(x)$ synthesized by the underlying deformation field
$\mathbf{u}(x)$.
Particularly, it proposed to 
minimize the following energy function 
\bq \label{eq:eng-func}
\min_u \;\; P(I_f, I_r; \mathbf{u}) \, + \, R(\mathbf{u}) \, ,
\eq
where $P(I_f, I_r; u)$ represents a dissimilarity measure of the two input
images $I_f(x)$ and $I_r(x)$ under the deformation field $\mathbf{u}(x) = (u_1(x), u_2(x), u_3(x))$ in 3D,
$R(\mathbf{u})$ is the convex regularization function to enforce deformations with
the required smoothness prior. 
In practice, the sum of absolute intensity differences (SAD) is often applied as a robust
dissimilarity measure of matching the input images $I_f(x)$ and $I_r(x)$ in
\eqref{eq:eng-func}:
\bq \label{eq:of-engd}                 
\min_u \; \; P(I_f, I_r; \mathbf{u}) \, := \, \int_{\Omega}
\abs{I_f(x+\mathbf{u}(x)) - I_r(x)}\, dx \, ;
\eq  
and the convex total-variation functions is employed as the
regularization term:
\bq \label{eq:of-lreg}
R(\mathbf{u}) \, := \, \alpha \, \sum_{i=1}^3 \, \int_{\Omega} \abs{\nabla u_i}  dx\, ,
\eq
which gives rise to a non-smooth energy function term in \eqref{eq:eng-func}.
Certainly, the smoothed function terms can also be studied in both image intensity matching and
smoothing deformations as in previous works~\cite{Horn81determiningoptical,Barron1994}.

Given the nonlinear functions of 
$I_r(x)$ and $I_f(x+ \mathbf{u}(x))$, the absolute intensity matching term
\eqref{eq:of-engd} in \eqref{eq:eng-func} is highly nonlinear and nonconvex in
general.
It is difficult to directly optimize the nonconvex energy function
\eqref{eq:eng-func}, even if the regularization term $R(\mathbf{u})$ is convex.
To address this issue efficiently, the sequential convex optimization
approach is introduced to minimize the proposed energy function \eqref{eq:eng-func} under a multi-scale
coarse-to-fine optimization perspective, where a series of linearization or convexification of the
nonlinear image intensity matching term
\eqref{eq:eng-func} at multiple image scales~\cite{yuan2007discrete} are contructed such that: 
first, we construct a coarse-to-fine pyramid of
each image function: let $I_f^1(x)$ \ldots $I_f^L(x)$ be the $L$-level
coarse-to-fine pyramid representation of the image $I_f(x)$ from the coarsest
resolution $I_f^1(x)$ to the finest resolution $I_f^L(x)=I_f(x)$; and $I_r^1(x)$
\ldots $I_r^L(x)$ the $L$-level coarse-to-fine pyramid representation of
the reference image $I_r(x)$.
At each $\ell$ level, $\ell=1 \ldots L$, we compute the deformation field
$\mathbf{u}^{\ell}(x)$ based on the two given image functions $I_r^{\ell}(x)$
and $I_f^{\ell}(x+\mathbf{u}^{\ell-1})$ at the same resolution level, where
$I_f^{\ell}(x+\mathbf{u}^{\ell-1})$ is warped by the deformation field
$\mathbf{u}^{\ell-1}(x)$ computed from the previous level $\ell-1$. For the
coarsest level, i.e. $\ell=1$, the initial previous-level deformation is set to be $0$.

To address the optimum deformation field $\mathbf{u}^*(x)$ of
\eqref{eq:eng-func}, let $\mathbf{u}^{\ell-1}(x)$ be the initial
estimation of $\mathbf{u}^*(x)$ at the current level $\ell$ for simplicity, $\mathcal{I}(x):=I_f(x + \mathbf{u}^{\ell-1}(x))$, and the
incremental deformation $\mathbf{h}(x)$ be the update of $\mathbf{u}^{\ell-1}(x)$ which
appropriately linearizes the image function $\mathcal{I}(x)$ over $\mathbf{h}(x)$ such that
\bq \label{eq:data-linear}
I_f\big((x + \mathbf{u}^{\ell-1}(x)) + \mathbf{h}(x)\big)  \approx  \mathcal{I}(x)\big(:=I_f(x + \mathbf{u}^{\ell-1}(x))\big) +
\nabla \mathcal{I}(x) \cdot \mathbf{h}(x) \, .
\eq

Given the image dissimilarity measure SAD \eqref{eq:of-engd}, 
the incremental deformation $\mathbf{h}(x)$ at each image scale $\ell$, $\ell=1 \ldots L$, can be
formulated such that
\bq \label{eq:of-la}
\min_{\mathbf{h}} \; \int_{\Omega} \abs{\mathcal{I}_0 + \nabla
\mathcal{I} \cdot \mathbf{h}} dx \, + \, R(\mathbf{u}^{\ell-1} + \mathbf{h}) \, ,
\eq
where  
\[
\mathcal{I}_0(x) := I_f(x + \mathbf{u}^{\ell-1}(x)) - I_r(x) \, ,
\quad \mathcal{I}(x):=I_f(x + \mathbf{u}^{\ell-1}(x))\, .
\]
This results in a sequence of
convex optimization problems, each of which properly
estimates the optimum update $\mathbf{h}^*(x)$ to the estimated deformation $\mathbf{u}^{\ell-1}(x)$ at the current image scale $\ell$,
then pass $\mathbf{u}^{\ell} = \mathbf{u}^{\ell-1} + \mathbf{h}^*$ to the update estimation at the next image scale $\ell + 1$ in sequence. 
%

For the resulted convex optimization problem \eqref{eq:of-la}, given the convex regularization term~\eqref{eq:of-lreg},  
we can derive its equivalent primal-dual and dual representations~\cite{sun2013efficient,rajchl2014rancor,sun2015a}, 
through a similar variational analysis as the primal-dual model \eqref{eq:pd-img-tv} and dual
model \eqref{eq:img-tv-dual} for the total-variation-based image processing problem~\eqref{eq:img-tv}, such that the corresponding primal-dual model
can be formulated as
\begin{align} \label{eq:of-pd}
\max_{w,q} \min_{\mathbf{h}} \; L(\mathbf{h}, w, q) := \int_{\Omega} \Big(w \mathcal{I}_0  \, + \,
\sum_{i=1}^3 u_i^{\ell-1} \, \D q_i \Big) dx  \, + \, \sum_{i=1}^3 \, \int_{\Omega} h_i \cdot F_i \, dx
\end{align}
subject to 
\bq \label{eq:of-dual-const2}
w(x) \leq 1 \, , \quad \abs{q_i(x)} \leq \alpha \, , \quad i \, = \, 1 \ldots 3 \, ,
\eq
where each $q_i$, $i=1 \ldots 3$, is the dual variable used for the total-variaion regularization term \eqref{eq:of-lreg}
in terms of \eqref{eq:dtvreg} and
\bq \label{eq:of-fdef}
F_i(x) \, := (w \cdot \partial_i \mathcal{I}  \,+ \, \D q_i)(x) \, , \quad i \, = \, 1 \ldots 3 \, .
\eq
The associate dual model, while minimizing the energy function $ L(\mathbf{h}, w, q)$ of \eqref{eq:of-pd} over each $h_i(x)$, $i=1 \ldots 3$, is
\bq \label{eq:of-dual}
\max_{w, q} \;  \int_{\Omega} \Big(w \mathcal{I}_0\,  + \,
\sum_{i=1}^3 u_i^{\ell-1} \, \D q_i \Big)\, dx 
\eq
subject to     
\bq \label{eq:of-dual-const}
F_i(x)  =  (w \cdot \partial_i \mathcal{I} \,  + \, \D q_i)(x) \,  = \, 0
\, , \quad i \, = \, 1 \ldots 3 \, ,
\eq   
and \eqref{eq:of-dual-const2}.

In view of the dual optimization problem \eqref{eq:of-dual}, the energy function $ L(\mathbf{h}, w, q)$ of \eqref{eq:of-pd} 
is clearly the Lagrangian function for which $h_i(x)$, $i=1 \ldots 3$, works as the multiplier to the 
linear equality constraint $F_i(x) = 0$ in \eqref{eq:of-dual-const}. To this end, we define the respective augmented Lagrangian 
function
\bq \label{eq:of-auglag}
L_c(\mathbf{h}, w, q) := \int_{\Omega} \Big(w \mathcal{I}_0  \, + \,
\sum_{i=1}^3 u_i^{\ell-1} \, \D q_i \Big) dx  \, + \, \sum_{i=1}^3 \, \int_{\Omega} h_i \cdot F_i \, dx
\, - \, \frac{c}{2} \, \sum_{i=1}^3 \, \norm{F_i(x)}^2 \, ,
\eq
and design the ALM-based non-rigid image registration algorithm Alg.~\ref{of-algd}

\begin{algorithm}[!h]
\caption{ALM-Based non-rigid image registration algorithm \label{of-algd}} 
Initialize $\mathbf{h}^0$ and $(w^0, q^0)$, for each iteration $k$ we explore the following two steps
\begin{itemize}
  \item fix $\mathbf{h}^k$, compute $(w^{k+1},q^{k+1})$:
  \bq \label{alg:of-stepp}
  (w^{k+1}, q^{k+1}) \, = \, \arg\max_{w,q} \, L_{c^k}(\mathbf{h}^k, p, q) \, , \quad \text{s.t.  \eqref{eq:of-dual-const2}}
  \eq
  provided the augmented Lagrangian function $L_c(\mathbf{h},w,q)$ in \eqref{eq:of-auglag};
  \item fix $(w^{k+1}, q^{k+1})$, then update $\mathbf{h}^{k+1}$ by
  \bq 
  h_i^{k+1} \, :=\, h_i^k - c^k \big(w^{k+1} \cdot \partial_i \mathcal{I} \,  + \, \D q_i^{k+1}\big) \, . \label{alg:of-stepu} 
  \eq
\end{itemize}
\end{algorithm}

\begin{figure}[!h]
\begin{center}
\includegraphics[height=5cm]{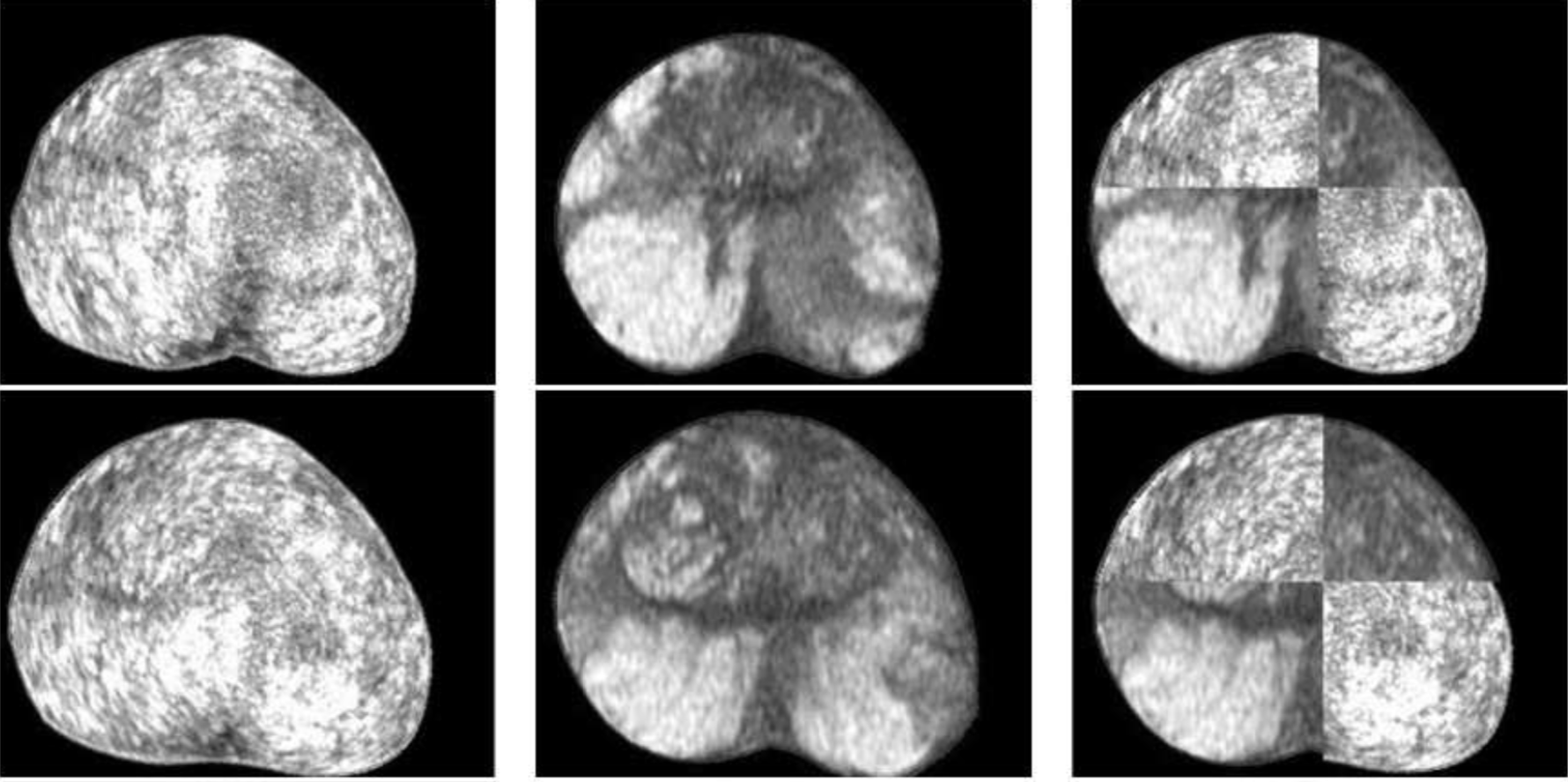}
 \end{center}
 \caption{\textit An experiment result of registering 3D prostate MRI to the fixed TRUS image, shown by selected
 slices~\cite{sun2013efficient,sun2015a}: (left) fixed TRUS image, (middle) registered MRI, (right) checkerboard
 of registration result computed by Alg.~\ref{of-algd}.
 \label{fig:of}}
 \end{figure}


The dual optimization-based algorithm Alg.~\ref{of-algd} properly avoids tackling the non-smooth function terms 
in the original optimization problem \eqref{eq:of-la} directly, and enjoys advantages in computing efficiency and robustness.
Experiment results~\cite{sun2013efficient,sun2015a} of 3D prostate MRI-TRUS registration showed such non-rigid image algorithm
achieved a high registration accuracy comparing to the conventional rigid registration, see Fig. \ref{fig:of} for illustration, where
a multi-channel modality independent neighborhood descriptor (MIND)~\cite{Heinrich20121423}, instead of gray-scale intensity information, was utilized
for such multi-modal image registration.

\subsection{Volume-Preserving Non-rigid Image Registration}

\begin{figure}[!h]
\begin{center}
\begin{tabular}[p]{c@{\,}c}
\subfigure[]{
  \includegraphics[height=3.6cm]{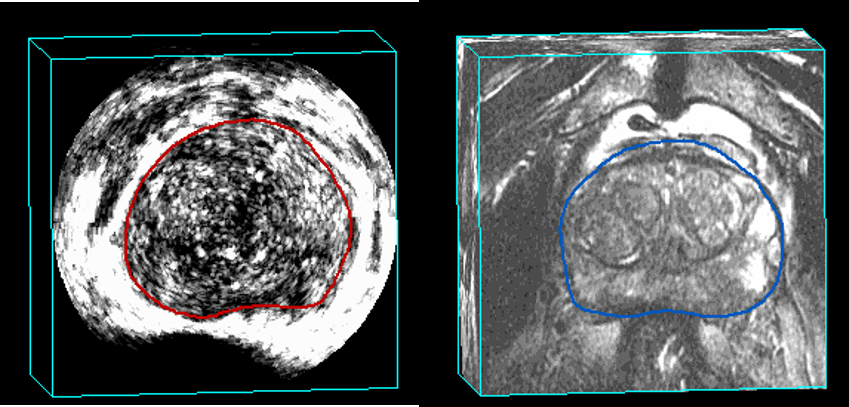}}&
  \subfigure[]{
  \includegraphics[height=3.6cm]{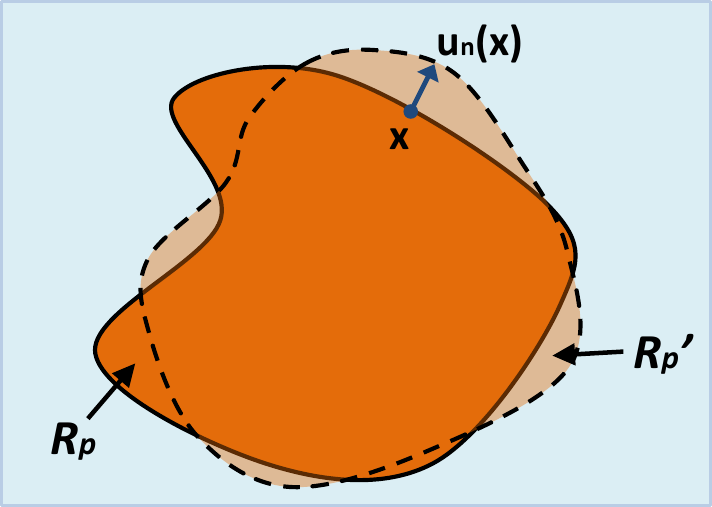}
  }
    \end{tabular}
 \end{center}
 \caption{\textit (a). An illustration of registered MR (right) - TRUS (left) images;
(b). the illustration of changes between the regions $R_p$ and $R_p^{'}$. 
 \label{fig:of-vp}}
\end{figure}

Non-rigid medical image registration is mostly challenging in practice. To improve its accuracy and reliability, more
prior information would be emplyed, for example, the volume-preserving prior w.r.t. a specific region~\cite{QY2016}, i.e. the volume of the
underlying region is expected to be preserved after registration, see Fig.~\ref{fig:of-vp} for demonstration.    

Given two input images $I_1(x)$ and $I_2(x)$, let $R_p$ be the specified region on the image $I_1(x)$, and $R_p^{'}$ be the
region of $R_p$ deformed over the deformation field $\mathbf{u}(x)$ on the second image $I_2(x)$, i.e. $R_p^{'} := R_p \circ \mathbf{u}$ (see Fig.~\ref{fig:of-vp} (b)).
We expect that the total volume change of the regions $R_p$ and $R_p^{'}$ is preserved or small enough, i.e. 
\bq \label{eq:vp_pr}
\delta V(\mathbf{u}) \, = \, Vol(R_p) - Vol(R_p^{'}(:=R_p \circ \mathbf{u})) \,.
\eq

Assume that the deformation field $\mathbf{u}(x)$ does not change the topology of the prostate region $R_p$, then the
volume change prior \eqref{eq:vp_pr} w.r.t.
$R_p$ results in 
\bq \label{eq:vp_pr02}
\delta V(u) \, = \,\int_{\Omega} \ell_{R_p} \cdot \D \mathbf{u} \, dx \, = \, \int_{\Omega} \nabla \ell_{R_p}
\cdot u \, dx \, ;
\eq
where $\ell_{R_p}(x)$ is the indicator function of the region $R_p$, i.e. $\ell_{R_p}(x)=1$ for $\forall x \in R_p$ and $\ell_{R_p}(x)=0$ otherwise.


In view of \eqref{eq:of-la}, we then formulate the volume-preserving non-rigid image registration problem at the image scale $\ell$ as
\bq \label{eq:of-vp}
\min_{\mathbf{h}} \; \int_{\Omega} \abs{\mathcal{I}_0 + \nabla
\mathcal{I} \cdot \mathbf{h}} dx \, + \, R(\mathbf{u}^{\ell-1} + \mathbf{h}) \, + \,  P(\delta {V}(u)) \, ,
\eq
where $P(\cdot) \geq 0$ is a convex penalty function, for which
$P(\delta {V}) = \gamma \abs{\delta {V}}$ or $P(\delta
{V}) = \chi(\delta {V} = 0)$, i.e. the linear equality constraint $\delta {V}(u) = 0$.

In fact, the divergence of the deformation field
$\mathbf{u}(x)$, i.e.
$\D \mathbf{u}(x) := (\partial_1 u_1 + \partial_2 u_2 + \partial_3 u_3)(x)$, represents the local volume change around each pixel $x \in R_p$; in consequence,
the lefthand side $\int_{\Omega} \ell_{R_p} \cdot \D \mathbf{u} \,dx$ in \eqref{eq:vp_pr02} exactly gives the total
accumulated volume change within the region $R_p$.
It is obvious that, for the incompressible deformation field within
the region $R_p$, i.e.
$
\D \mathbf{u}(x)  =  0$ $\forall x  \in R_p
$,
which means the local volume change at each pixel $x \in R_p$ exactly vanishes; the total volume change $\delta
{V}$ is therefore preserved and the volume preserving prior \eqref{eq:vp_pr02} is definitely satisfied. 
Clearly, such
incompressible condition is only the special case of \eqref{eq:vp_pr02} and over-constrains the desired
deformation field since it does not allow any local volume change; in contrast, the proposed volume preserving prior
$P(\delta {V})$, even for the exact linear equality constraint $\delta {V}(\mathbf{u}) = 0$, does allow the local
volume change within the region $R_p$.

For the resulted convex optimization problem \eqref{eq:of-vp}, 
we can apply a similar variational analysis strategy as \eqref{eq:of-pd} and \eqref{eq:of-dual} to build equivalent optimization models,
and derive the identical primal-dual model to \eqref{eq:of-vp} as
\begin{align} \label{eq:of-vp-pd}
\hspace{-0.3cm}
\max_{w,q} \min_{\mathbf{h}} \; L(\mathbf{h}, w, q, \pi) := \int_{\Omega} \Big(w \mathcal{I}_0  \, + \,
\sum_{i=1}^3 u_i^{\ell-1} \, \D q_i \, + \,  \pi \nabla \ell_{R_p} \cdot \mathbf{u} \Big) dx  \, + \, \sum_{i=1}^3 \, \int_{\Omega} h_i \cdot G_i \, dx
\end{align}
subject to
\bq \label{eq:of-vp-dual-const2}
w(x) \leq 1 \, , \quad \abs{q_i(x)} \leq \alpha \, , \quad i \, = \, 1 \ldots 3 \,; \quad \abs{\pi} \leq \gamma \, 
\eq
where the functions $G_i(x)$, $i=1 \ldots 3$, are defined as
\bq \label{eq:of-vp-fdef}
G_i(x) \, := \,(w \cdot \partial_i \mathcal{I}  \,+ \, \D q_i \, + \,  \pi\,
\partial_i \ell_{R_p})(x)\, , \quad i \, = \, 1 \ldots 3 \, .
\eq
Minimizing the energy function $ L(\mathbf{h}, w, q, \pi)$ of \eqref{eq:of-vp-pd} over each $h_i(x)$, $i=1 \ldots 3$, first, we obtain
the associate dual model:
\bq \label{eq:of-vp-dual}
\max_{w, q, \pi} \;  \int_{\Omega} \Big(w \mathcal{I}_0  \, + \,
\sum_{i=1}^3 u_i^{\ell-1} \, \D q_i \, + \,  \pi \nabla \ell_{R_p} \cdot \mathbf{u} \Big) dx
\eq
subject to     
\bq \label{eq:of-vp-dual-const}
G_i(x)\,  = \, (w \cdot \partial_i \mathcal{I}  \,+ \, \D q_i \, + \,  \pi\,
\partial_i \ell_{R_p})(x) \,  = \, 0
\, , \quad i \, = \, 1 \ldots 3 \, ,
\eq   
and the constraints \eqref{eq:of-vp-dual-const2} on the variables $w(x)$, $q_i(x)$, $i=1\ldots 3$, and $\pi$.

In terms of the dual optimization model \eqref{eq:of-vp-dual}, the energy function $L(\mathbf{h}, w, q, \pi)$ of \eqref{eq:of-vp-pd} 
just works as the Lagrangian function, where $h_i(x)$, $i=1 \ldots 3$, is the multiplier to the 
linear equality constraint $G_i(x) = 0$ of \eqref{eq:of-vp-dual-const}. Hence, the respective augmented Lagrangian 
function can be given as
\bq \label{eq:of-vp-auglag}
L_c(\mathbf{h}, w, q, \pi) \,:= \, L(\mathbf{h}, w, q, \pi)
\, - \, \frac{c}{2} \, \sum_{i=1}^3 \, \norm{G_i(x)}^2 \, ,
\eq
and design the ALM-based non-rigid image registration algorithm Alg.~\ref{of-vp-algd}.

\begin{algorithm}[!h]
\caption{ALM-Based non-rigid image registration algorithm \label{of-vp-algd}} 
Initialize $\mathbf{h}^0$ and $(w^0, q^0, \pi^0)$, for each iteration $k$ we explore the following two steps
\begin{itemize}
  \item fix $\mathbf{h}^k$, compute $(w^{k+1},q^{k+1}, \pi^{k+1})$:
  \bq \label{alg:ofvp-stepp}
  (w^{k+1}, q^{k+1}, \pi^{k+1}) \, = \, \arg\max_{w,q, pi} \, L_{c^k}(\mathbf{h}^k, p, q, \pi) \, ,
  \quad \text{s.t.  \eqref{eq:of-vp-dual-const2}}
  \eq
  provided the augmented Lagrangian function $L_c(\mathbf{h},w,q, \pi)$ in \eqref{eq:of-vp-auglag};
  \item fix $(w^{k+1}, q^{k+1}, \pi^{k+1})$, then update $\mathbf{h}^{k+1}$ by
  \bq 
  h_i^{k+1} \, :=\, h_i^k - c^k \big(w^{k+1} \cdot \partial_i \mathcal{I} \,  + \, \D q_i^{k+1} \, + \, 
  \pi\, \partial_i \ell_{R_p}\big) \, . \label{alg:ofvp-stepu} 
  \eq
\end{itemize}
\end{algorithm}

In experiments, the penalty parameter $\gamma >0$ in $P(\delta {V}) = \gamma \abs{\delta {V}}$
can take a pretty big value, this approximates the exact volume-preserving prior.
Experiment results of 3D prostate MR-TRUS registration showed that the registration accuracy is 
significantly improved from $83.5\pm 3.8 \%$ (by the non-rigid registration approach \eqref{eq:of-la}) to $87.3 \pm 3.4 \%$ 
(by the volume-preserving registration method \eqref{eq:of-vp}) in DSC, see \cite{QY2016} for details.

\subsection{Spatial-Temporal Non-rigid Registration of Medical Images}

Some clinical-based image analysis tasks often require registering a sequence of images,  
the acquired images at different time-spots are aligned sequentially to
quantatively monitor temporal developments of the studied biomarkers or evaluating treatments.
For example, the sequence of 3D ultrasound images of the pre-term newborn's brain can be 
developed to monitor the ventricle volume as a biomarker for longitudinally analyzing ventricular dilatation and deformation.
This allows for the precise analysis of local  
ventricular changes which could affect specific white matter bundles, 
such as in the motor or visual cortex, and could be linked to specific neurological 
problems often seen in this patient population later in life \cite{qy2015miccai,qiuyj2017}. 

Given a sequence of images $I_1(x)$ \ldots $I_{n+1}(x)$, we aim to compute the temporal sequence of 
3D non-rigid deformation fields $u_k(x)$,
$k=1 \ldots n$, within each two consecutive images $I_k(x)$ and $I_{k+1}(x)$, 
while imposing both spatial and temporal smoothness 
of the {spatial-temporal} deformation fields $u_k(x) = (u_k^1(x), u_k^2(x), u_k^3(x))^\tT$,
$k=1\ldots n$.

Besides the spatial-smoothness-regularized deformation estimation within two sequential images $I_k(x)$ and $I_{k+1}(x)$, $k=1 \ldots n$, through
solving a series of optimization problems \eqref{eq:of-la},
the additional temporal smoothness prior encourages the similarities between each
two consecutive deformation fields $u_k(x)$ and $u_{k+1}(x)$, $k=1 \ldots n-1$, for example, penalizes their total absolute
differences
\bq \label{eq:stof-term3}
T(\mathbf{u}) \, := \, \gamma\, \sum_{k=1}^{n-1} \int_{\Omega} \big(\abs{u_k^1 -
u_{k+1}^1} + \abs{u_k^2 - u_{k+1}^2} + \abs{u_k^3 -
u_{k+1}^3}\big)\, dx \, ,
\eq
where $\gamma >0$ is the temporal regularization parameter. 
Such absolute function-based proposed
temporal regularization function \eqref{eq:stof-term3} can significantly eliminate the undesired sudden changes within each
two deformation fields, which is mainly due to the poor image quality of US including strong US
speckles and shadows, low tissue contrast, fewer image details of structures, 
and improve robustness of the introduced spatial-temporal non-rigid registration method.

Also, under the multi-level image registration framework, the \emph{spatial-temporal} non-rigid image registration estimate
the update deformation field $\mathbf{h}_k(x)=(h_{k,1}, h_{k,2}, h_{k,3})(x)$, $k=1 \ldots n$, at the image scale level $k$through the following convex optimization problem
\bq \label{eq:stof}
\min_{\mathbf{h}} \; \sum_{k=1}^n \,\int_{\Omega} \abs{\mathcal{I}_0^k + \nabla
\mathcal{I}_k \cdot \mathbf{h}_k} dx \, + \, \sum_{k=1}^n R(\mathbf{u}_k^{\ell-1} + \mathbf{h}_k)
\, + \, T(\mathbf{u}^{\ell-1} + \mathbf{h}) \, ,
\eq
where the tempopral regularization term $T(\cdot)$ is given in \eqref{eq:stof-term3} and
\[
\mathcal{I}_0^k(x) := I_{k}(x + \mathbf{u}_k^{\ell-1}(x)) - I_{k+1}(x) \, ,
\quad \mathcal{I}_k(x):=I_k(x + \mathbf{u}_k^{\ell-1}(x)) \, ,
\]
and $\mathbf{u}_k^{\ell-1}(x)=(u_{k,1}^{\ell-1}, u_{k,2}^{\ell-1}, u_{k,3}^{\ell-1})(x)$, $k=1 \ldots n$.

For the studied convex optimization problem \eqref{eq:stof}, 
the similar variational analysis, as the equivalent models \eqref{eq:of-pd} and \eqref{eq:of-dual} to \eqref{eq:of-la}, can be explored to obtain
its identical primal-dual model, see \cite{qy2015miccai,qiuyj2017} for more details, such that
\begin{align}
\max_{w,q,r} \min_{\mathbf{h}} \; L(\mathbf{h}, w, q, r) \, :=& \, \sum_{k=1}^n \int_{\Omega} \Big(w_k \mathcal{I}_0^k  \, + \,
\sum_{i=1}^3 u_{k,i}^{\ell-1} \, \D q_{k,i} \Big) dx  \, \, + \label{eq:stof-pd}\\
& \,\sum_{k=1}^{n} \sum_{i=1}^3 \, \int_{\Omega} h_{k,i} \cdot F_{k,i} \, dx \, + \, \sum_{k=1}^{n-1} \sum_{i=1}^3\int_{\Omega} \big(r_{k,i} (u_{k,i}^{\ell-1} -
u_{k+1, i}^{\ell-1}) \big) \, dx \, \nonumber
\end{align}  
subject to
\bq \label{eq:stof-const}
\abs{w_k(x)} \, \leq \, 1 \, , \quad \abs{r_{k,i}(x)} \, \leq \, \gamma \, ,  \quad \abs{q_{k,i}(x)} \leq \alpha \, , \quad k = 1
\ldots n\, , \; i \, = \, 1 \ldots 3 \, ;
\eq
where
\begin{align} \label{eq:stof-dconst1}
F_{1,i}(x) \, :=  &(w_1 \cdot \partial_i \mathcal{I}_1  +  \D q_{1,i} + r_{1,i})(x)\,
  , \quad i\, = \, 1 \ldots 3\\ \label{eq:stof-dconst2}
F_{k,i}(x) \, :=  &(w_k \cdot \partial_i \mathcal{I}_k  +  \D q_{k,i} +
(r_{k,i} - r_{k-1,i}))(x)\, , \;\; \, k = 2 \ldots n-1\, , \; i = 1 \ldots 3
\\ \label{eq:stof-dconst3}
F_{n,i}(x) \, :=  &(w_n \cdot \partial_i \mathcal{I}_n  +  \D q_{n,i} -
r_{n-1,i})(x)\, , \quad i\, = \, 1 \ldots 3 \, .
\end{align}

Minimizing the energy function $ L(\mathbf{h}, w, q, r)$ of \eqref{eq:stof-pd} over each $h_{k,i}(x)$, $k=1 \ldots n$ and $i=1 \ldots 3$, we obtain
the corresponding dual model:
\bq \label{eq:stof-dual}
\max_{w, q, r} \;  \sum_{k=1}^n \int_{\Omega} \Big(w_k \mathcal{I}_0^k  \, + \,
\sum_{i=1}^3 u_{k,i}^{\ell-1} \, \D q_{k,i} \Big) \, dx  \, + 
 \, \sum_{k=1}^{n-1} \sum_{i=1}^3\int_{\Omega} \big(r_{k,i} (u_{k,i}^{\ell-1} -
u_{k+1, i}^{\ell-1}) \big) \, dx \, 
\eq
subject to the constraints \eqref{eq:stof-const} and the linear equality constraints
\bq \label{eq:stof-lq-const}
F_{k,1}(x)\,  = \,  0
\, , \quad k= 1\ldots n \, , \; i  = 1 \ldots 3 \, ,
\eq   
for the given linear functions \eqref{eq:stof-dconst1} - \eqref{eq:stof-dconst3}.

Observing the dual optimization model \eqref{eq:stof-dual}, the function $L(\mathbf{h}, w, q, r)$ of \eqref{eq:stof-pd} 
just works as the Lagrangian function, where each $h_{k,i}(x)$, $k=1\ldots n$ and $i=1 \ldots 3$, is the multiplier to the 
linear equality constraint $F_{k,i}(x) = 0$ of \eqref{eq:stof-lq-const}. We define the respective augmented Lagrangian 
function
\bq \label{eq:stof-auglag}
L_c(\mathbf{h}, w, q, r) \,:= \, L(\mathbf{h}, w, q, r)
\, - \, \frac{c}{2} \,\sum_{k=1}^n \sum_{i=1}^3 \, \norm{F_{k,i}(x)}^2 \, ,
\eq
and the ALM-based non-rigid image registration algorithm can be formulated as Alg.~\ref{stof-algd}.

\begin{algorithm}[!h]
\caption{ALM-Based non-rigid image registration algorithm \label{stof-algd}} 
Initialize $\mathbf{h}^0$ and $(w^0, q^0, r^0)$, for each iteration $j$ we explore the following two steps
\begin{itemize}
  \item fix $\mathbf{h}^j$, compute $(w^{j+1},q^{j+1}, r^{j+1})$:
  \bq \label{alg:stof-stepp}
  (w^{j+1}, q^{j+1}, r^{j+1}) \, := \, \arg\max_{w,q, r} \, L_{c^j}(\mathbf{h}^j, p, q, r) \, ,
  \quad \text{s.t.  \eqref{eq:stof-const}}
  \eq
  provided the augmented Lagrangian function $L_c(\mathbf{h},w,q, r)$ in \eqref{eq:stof-auglag};
  \item fix $(w^{j+1}, q^{j+1}, r^{j+1})$, then update $\mathbf{h}^{j+1}$ by
  \begin{align*}
  h_{1,i}^{j+1} \, = & \, h_{1,i}^j - c^j \big(w_1^{j+1} \cdot \partial_i \mathcal{I}_1  +  \D q_{1,i}^{j+1} + r_{1,i}^{j+1}\big) \, , \quad i =  1 \ldots 3\\
  h_{k,i}^{j+1} \, = &\, h_{k,i}^{j} - c^j \big( w_k^{j+1} \cdot \partial_i \mathcal{I}_k  +  \D q_{k,i}^{j+1} +
(r_{k,i}^{j+1} - r_{k-1,i}^{j+1})\big) \, , \;\; \, k = 2 \ldots n-1\, , \; i = 1 \ldots 3 \\
h_{n,i}^{j+1} \, = & \, h_{n,i}^{j} - c^j \big( w_n^{j+1} \cdot \partial_i \mathcal{I}_k  +  \D q_{n,i}^{j+1} -
r_{n-1,i}^{j+1}\big) \, \quad i\, = \, 1 \ldots 3 \, .
  \end{align*}
\end{itemize}
\end{algorithm}

\begin{figure}[!h]
\begin{center}
\begin{tabular}[p]{c@{\,}c}
\subfigure[]{
  \includegraphics[height=4.38cm]{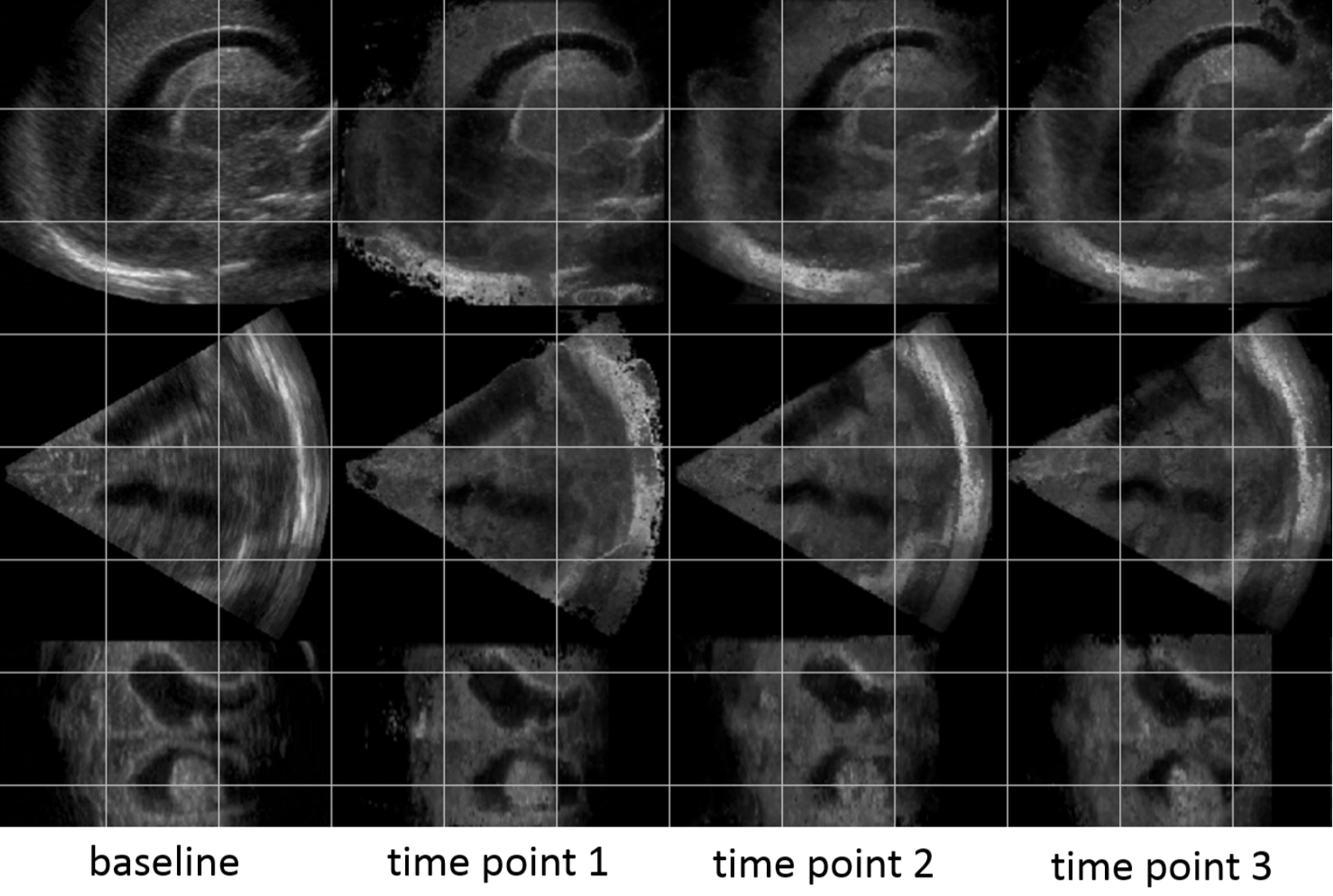}}&
  \subfigure[]{
  \includegraphics[height=4.38cm]{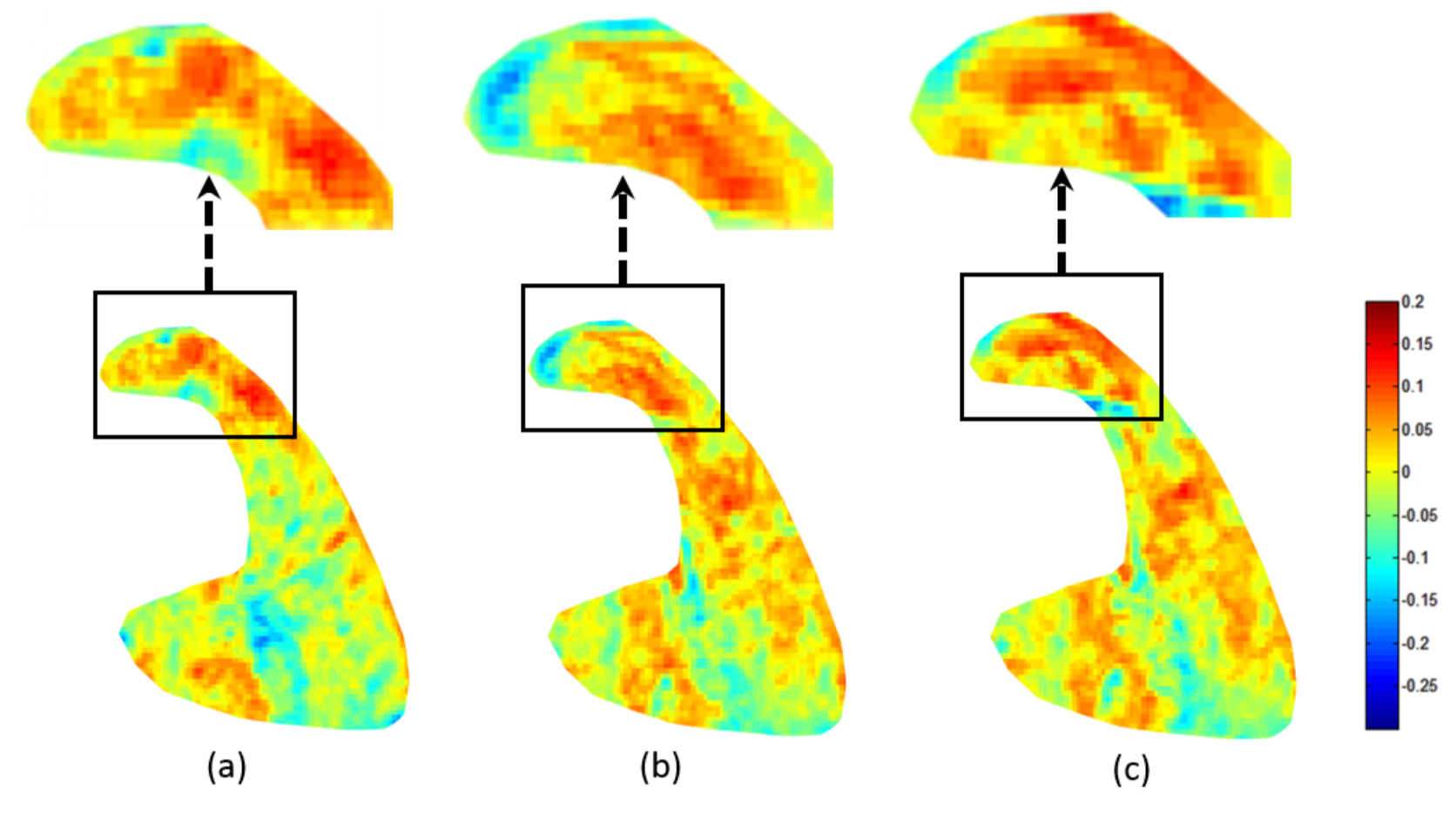}
  }
    \end{tabular}
 \end{center}
 \caption{\textit (a). Registration results of IVH neonatal ventricles 
 in a 3D US image sequence \cite{qy2015miccai,qiuyj2017}; $1$st - $4$th columns provide baseline, registered images at time 
 points $1$ - $3$; $1$st - $3$rd row: saggital, coronal, and transvers view.
(b). Local volume changes represented by the divergence of deformation, 
i.e. $\D \mathbf{u}(x)$, at each time point; a - c: the first-third time points,
where the local volume expansion, i.e. $\D \mathbf{u}(x) > 0$, is colored in red, while the 
local volume shrinkage, i.e. $\D \mathbf{u}(x)< 0$, is colored in blue;
the regions inside the black frames are zoomed and shown in images on the top row.
 \label{fig:stof}}
\end{figure}

Experiment results of registering IVH neonatal ventricles 
in a 3D US image sequence is demonstrated in Fig. \ref{fig:stof}, which
shows most important information about local volume changes at different time spots
to help clinicians monitoring ventricle developments and evaluating 
treatments \cite{qy2015miccai,qiuyj2017}.
%
%